\documentclass[12pt,times]{article}
\title{Deep neural operator for learning transient response of interpenetrating phase composites subject to dynamic loading}

\author{Minglei Lu$^1$, Ali Mohammadi$^1$, Zhaoxu Meng$^1$, Xuhui Meng$^2$,\\ Gang Li$^1$ and Zhen Li$^1$\footnote{Email: \href{mailto:zli7@clemson.edu}{zli7@clemson.edu}}\\
\small{$^1$ Department of Mechanical Engineering, Clemson University, Clemson, SC 29634, USA}\\
\small{$^2$ School of Mathematics and Statistics, Huazhong University of Science and Technology,}\\
\small{Wuhan, Hubei 430074, China} }

\date{\today}

\RequirePackage{lineno}
\usepackage[colorlinks=true]{hyperref}
\usepackage{graphicx}
\usepackage{multirow,array,color,mathrsfs,subcaption}
\usepackage{epstopdf} 
\usepackage{booktabs}
\usepackage{caption}
\DeclareGraphicsExtensions{.eps,.mps,.pdf,.jpg,.png}

\usepackage{fancyhdr}

\usepackage{amsmath}
\usepackage{amsfonts}
\usepackage[sort&compress,numbers]{natbib}
\usepackage{dsfont}

\usepackage[top=1.5cm,bottom=2.5cm,left=1.5cm,right=1.5cm]{geometry}

\usepackage[figuresright]{rotating}
\usepackage{extarrows}

\usepackage{fixltx2e}
\usepackage{xspace}
\usepackage{amsmath,amssymb}

\begin{document}

\maketitle

\begin{abstract}
Additive manufacturing has been recognized as an industrial technological revolution for manufacturing, which allows fabrication of materials with complex three-dimensional (3D) structures directly from computer-aided design models. Using two or more constituent materials with different physical and mechanical properties, it becomes possible to construct interpenetrating phase composites (IPCs) with 3D interconnected structures to provide superior mechanical properties as compared to the conventional reinforced composites with discrete particles or fibers. The mechanical properties of IPCs, especially response to dynamic loading, highly depend on their 3D structures. In general, for each specified structural design, it could take hours or days to perform either finite element analysis (FEA) or experiments to test the mechanical response of IPCs to a given dynamic load. To accelerate the physics-based prediction of mechanical properties of IPCs for various structural designs, we employ a deep neural operator (DNO) to learn the transient response of IPCs under dynamic loading as surrogate of physics-based FEA models. We consider a 3D IPC beam formed by two metals with a ratio of Young's modulus of 2.7, wherein random blocks of constituent materials are used to demonstrate the generality and robustness of the DNO model. To obtain FEA results of IPC properties, 5,000 random time-dependent strain loads generated by a Gaussian process kennel are applied to the 3D IPC beam, and the reaction forces and stress fields inside the IPC beam under various loading are collected. Subsequently, the DNO model is trained using an incremental learning method with sequence-to-sequence training implemented in JAX, leading to a 100X speedup compared to widely used vanilla deep operator network models. After an offline training, the DNO model can act as surrogate of physics-based FEA to predict the transient mechanical response in terms of reaction force and stress distribution of the IPCs to various strain loads in one second at an accuracy of 98\%. Also, the learned operator is able to provide extended prediction of the IPC beam subject to longer random strain loads at a reasonably well accuracy. Such superfast and accurate prediction of mechanical properties of IPCs could significantly accelerate the IPC structural design and related composite designs for desired mechanical properties.
\end{abstract}

\section{Introduction}\label{sec1}
\vspace{-5pt}
There are numerous examples of composite materials in nature, including human bone and teeth, wood, pearls, and shell structures~\cite{2015-Murr-Examples}. A composite material is made by combining two or more constituent materials with notably different physical and mechanical properties to have physio-thermo-mechanical properties distinct from those of its individual components~\cite{2022-Tamura-Origin}. Though composites are made of single-phase materials, the physical and mechanical properties of composites are not simple linear combinations of those properties of each component. With appropriate arrangement of the single-component elements, composite materials can have superior functions such as high strength-to-weight ratio, high fracture resistance, and resistance to corrosion and fatigue exceeding the simple combinations of materials properties~\cite{2014-Zheng-Ultralight,2015-Clausen-Topology}. Because composite materials can be easily found in nature and biology, such as pearl shells with a lamellar structure of hard layers for fracture resistance and soft layers for energy absorption~\cite{2009-Aoyanagi-Stress}, and natural wood with hierarchical structures of long cellulose fibres held together by lignin~\cite{2020-Solala-OnThe}, great efforts have been made to create composite materials by mimicking natural composites and bio-materials~\cite{2021-Ghazlan-Inspiration} for practical uses in airplanes and vehicles~\cite{2022-Tiwary-AReview}, which have actively promoted the development of novel composite materials with expected nonlinear physical and mechanical properties.

The mechanical properties of composites, especially the response to dynamic loading, highly depend on their internal structures that determine how distinctive single-phase elements are arranged. In general, optimizing the topology of composite materials to design optimal microstructure of material phases can lead to better multi-functional composites~\cite{2022-Ma-Strength}. However, complex geometries and three-dimensional (3D) microstructures of composites are very difficult to fabricate using traditional manufacturing methods. Therefore, most widely used composite materials have discrete particles or fibers as the enforcing phase embedded in the matrix phase~\cite{2018-Brenken-Fused}. Enabled by the development of revolutionary additive manufacturing (AM) or 3D printing technique~\cite{2018-Moustafa-Mesostructure}, it becomes possible to construct interpenetrating phase composites (IPCs) with 3D interconnected structures to provide superior mechanical properties as compared to the conventional reinforced composites with discrete particles or fibers.

AM provides great freedom to design complex 3D geometry, material combination, and multi-functional properties~\cite{2016-Thompson-design}, making it an important methodology in modern manufacturing. Taking the advantages of its flexibility and ability to process a wide range of materials such as metals, polymers, and ceramic materials~\cite{1996-W-microstr,2003-San-alumina}, AM provides the means for computer-aided design and manufacturing to construct complex 3D structures.
Most existing composite materials consist of discrete fibres or particles which are dispersed in a binding matrix phase~\cite{2019-Goh-Recent}. In contrast, IPC contains no isolated phases, which means that if any one of the constituent phases were removed, the remaining phases would still form a self-supporting, open-celled foam~\cite{2000-LDW-mechanical1}. Compared to traditional discontinuously reinforced composite materials, such independently self-supporting and load-bearing features of IPC have been investigated~\cite{2019-Bonatti-mechanical}.
Recent advancements in AM have made it possible to fabricate IPCs with controlled and complex topologies~\cite{2017-Palaganas-3D,2017-Wang-3D,2021-Zhang-mechanical}. As a result, a large number of IPCs with different structures and materials were created, making the investigation of IPC properties an indispensable study~\cite{1995-Helge-strength,2008-Poniznik-effective,2014-Cheng-modeling,2017-Al-mechanical}. Many efforts on experiments~\cite{2009-Binner-processing,2015-Liu-cyclic,2017-Okulov-dealloying,2019-Liu-mechanical} and numerical computations~\cite{2009-Jhaver-processing,2014-Li-simulation} have been made to quantify the mechanical properties of IPCs in different applications. As one of the important aspects to evaluate, the transient response can reflect many mechanical properties of composites under dynamic loading. In general, IPCs are prone to fracture and failure along the loading axis under high-strain loading~\cite{2014-Wang-damage,2000-LDW-mechanical2}. Therefore, it is of great importance to study the transient response of IPCs to dynamic loading.

Given an IPC object with complex 3D structures, numerical methods can be used to study the transient response of the IPC object, where finite element analysis (FEA) can discretize large IPC systems into small elements and solve corresponding partial differential equations (PDEs) on these elements. FEA is a well-established computational method to solve PDEs and has been widely used to study the behavior of different kinds of composites~\cite{2022-Li-ceramic,2022-Tong-experiment, 2021-Guo-constitutive, 2021-Metin-in-plane}.
However, identifying and formulating fundamental PDEs suitable for modeling a particular problem requires extensive prior domain knowledge of the corresponding field. In some cases, it can be tough and time-consuming to perform numerical simulations, especially for time-dependent problems.
Furthermore, even if a particular composite material is well studied, it still requires experimental or numerical work to obtain nonlinear mechanical responses in terms of stress and displacement fields when it is subjected to different dynamic loads. It means that any new design of composite materials needs to wait hours or days to know the material properties, which could significantly slowdown the material design and optimization process.
With the continuous improvement of computing power, machine learning is gradually more widely integrated into complex analysis and prediction in various research fields.
Introducing machine learning models as surrogate of physics-based FEA models or experiments to learn the transient mechanical response of composites to dynamic loading and predict material properties at no time could be an efficient way to accelerate the design process of composites.

Different deep learning approaches have been developed and successfully applied to diverse physical problems in recent years~\cite{2021-Lin-Operator, 2022-Malidarre-invest, 2022-Cai-applic, 2022-Yin-interf, 2022-Tao-finite}, wherein deep neural operator (DNO) is a deep learning framework to learn effective nonlinear operators mapping between infinite dimensional function spaces~\cite{2021-kovachki-neural}. Deep Operator Network (DeepONet), as one type of DNO, brings a new solution for finding such nonlinear operators~\cite{2020-Li-Fourier}. DeepONet can learn continuous nonlinear operators between input and output~\cite{2021-Lu-learning}, so that it can be used to approximate various explicit and implicit mapping functions like Laplace transform and PDEs, which are the most common but difficult mathematical relationships to investigate in various dynamic systems.
To find effective nonlinear operators mapping between two time-dependent functions, the traditional DeepONet is trained on data of time sequence point-by-point. When it encounters long time sequences, the training process could be extremely time-consuming. To this end, we will develop and implement an sequence-to-sequence training method for the DNO models to improve the DNO training efficiency in long time-dependent problems.

\begin{figure}[t]
    \centering
    \includegraphics[width=0.49\textwidth]{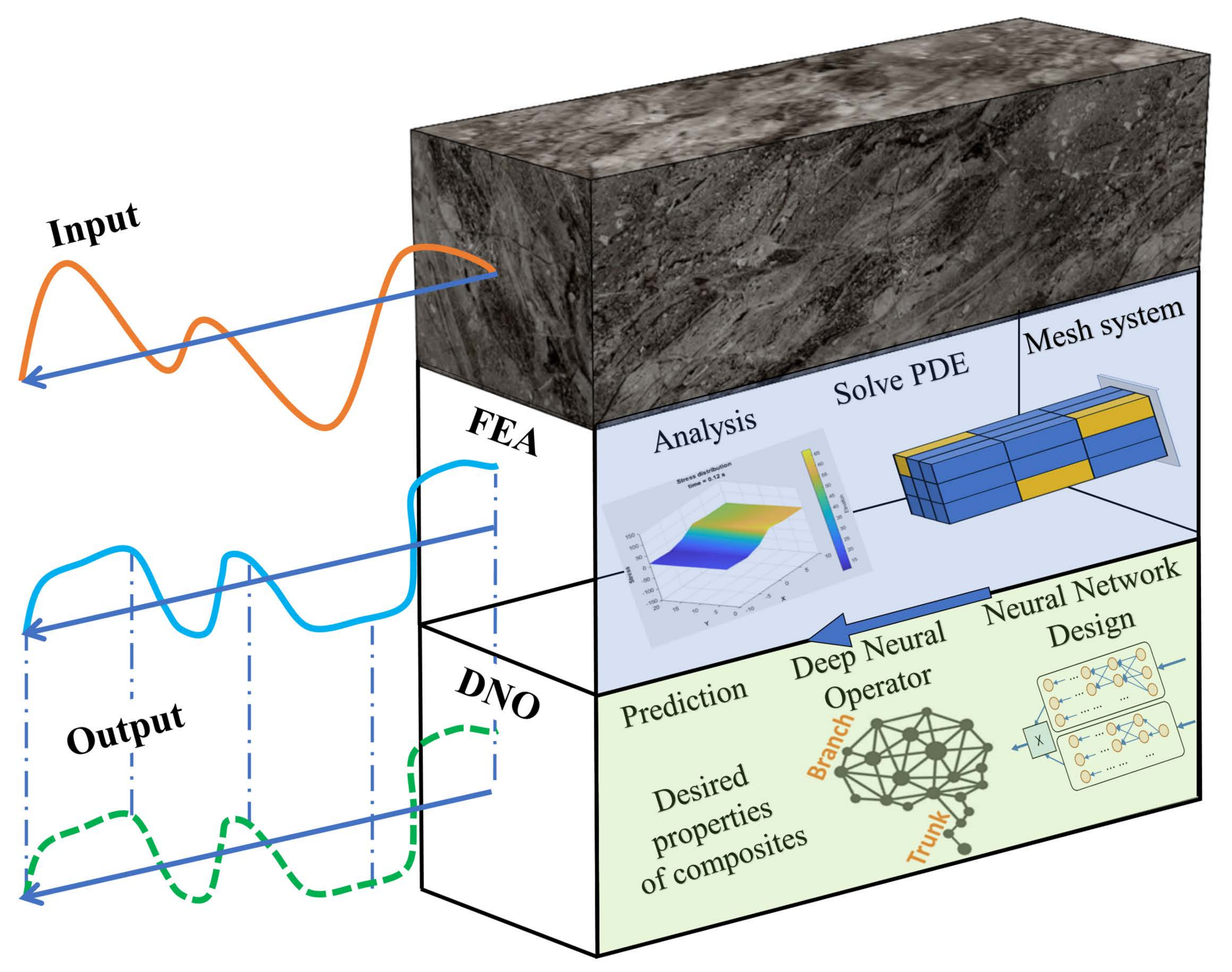}
    \caption{\textbf{Schematic of composites analysis with deep neural operator.} At the top layer, the composites to be studied is selected. At the middle layer, FEA is applied to analyze the composites and collect the data. At the bottom layer, the proposed model can be trained by the collected data. And then, the trained model can predict the results corresponding to arbitrary inputs.}
    \label{fig1}
\end{figure}

In this paper, we consider training a DNO model to act as surrogate of physics-based models of an IPC beam to provide fast and accurate predictions of the transient mechanical responses of this IPC beam to dynamic loading. Fig.~\ref{fig1} illustrates the learning framework, where the physical system of an IPC beam will be created and simulated using FEA, providing mechanical responses of the IPC beam to different dynamic loads as training data for the DNO model. An improved DNO model will be developed based on the DeepONet framework to enable sequence-to-sequence training. We shall demonstrate the effectiveness of the DNO model and investigate its prediction accuracy, the training efficiency, generalizability to different inputs, and robustness to noise.

The remainder of this paper is organized as follows. Section~\ref{sec2} introduces the problem setup, an improved deep operator learning framework with sequence-to-sequence learning, and an incremental training algorithm to further improve the training efficiency. Section~\ref{sec3} presents the computational results showing the performance of the proposed DNO model for predicting transient responses of an IPC beam subject to dynamic loading. Also, we investigate the effectiveness, extensibility, and robustness of the proposed DNO model. Finally, Section~\ref{sec4} concludes the main findings of this work with a brief summary and discussion.

\section{Methods}\label{sec2}
We consider a 3D IPC beam subject to any arbitrary dynamic loading as an example system to show the effectiveness of the proposed DNO model and its implementation, where the IPC bean is made of random blocks of constituent materials to demonstrate the generality and robustness of the DNO model. It is worth emphasizing that the DNO model and improved training algorithms are not limited to this IPC beam system, and are readily applied to related composites for superfast and accurate prediction of nonlinear mechanical properties.
In this section, we will first introduce the problems setup, and then the details of the DNO model and an incremental training algorithm.

\subsection{Problem setup}\label{sec2.1}
\begin{figure}[t]
    \centering
    \includegraphics[width=0.49\textwidth]{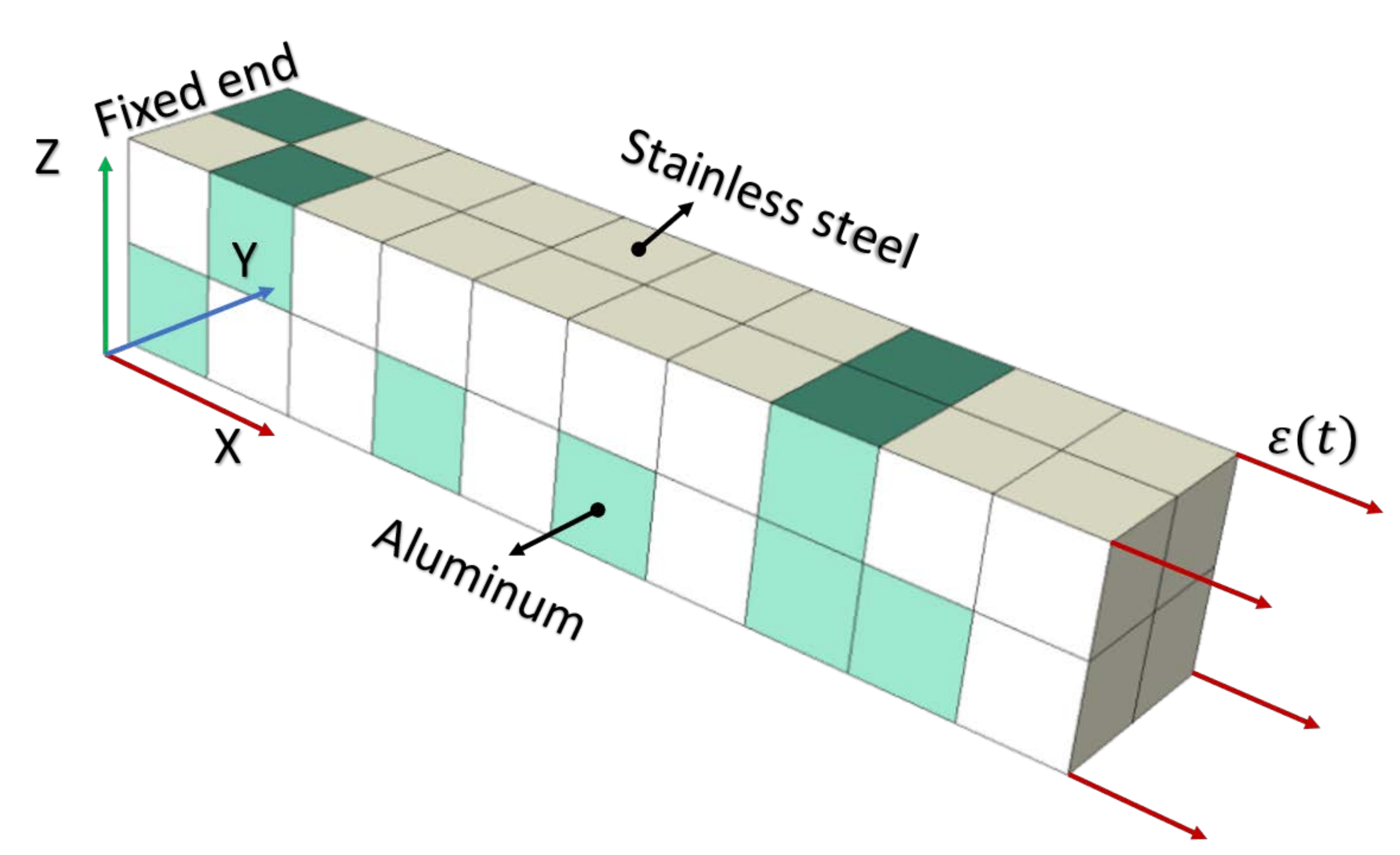}
    \caption{\textbf{Computational domain and boundary conditions}. The cuboid composite beam of size $500~{\rm mm}\times100~{\rm mm}\times100~{\rm mm}$ is divided into 40 identical cubic subdomains assigned with either stainless steel or aluminum alloy, with a light color (white) repenting 27 stainless steel elements and dark color (green) being 13 aluminum alloy elements. One end of the beam is fixed, and the other end is free and is imposed an arbitrarily dynamic strain loading.}
    \label{fig2}
\end{figure}

An IPC beam with a cuboid structure composed of stainless steel and aluminum alloy is created, with one fixed end and one free end, where an arbitrarily dynamic strain loading is applied to the free end, as shown in Fig.~\ref{fig2}.
The IPC beam is set to $500~{\rm mm}$ in length with a cross section of $100~{\rm mm}\times100~{\rm mm}$. The structure is divided into 40 identical cubic subdomains, each with a side length of $50~{\rm mm}$.
Stainless steel and aluminum alloy are randomly assigned to these cubes to form a composite beam, with details of material properties listed in Table~\ref{tab1}.
Taking into account both the elastic and plastic properties of the stainless steel and the aluminum alloy with significantly different elastic modulus and tensile strengths, the composite beam becomes highly nonlinear under dynamic loading. The elasticity of the material is determined by Young's modulus and Poisson's ratio, while the plasticity of the material is described by a multilinear isotropic hardening model based on available stress-strain curves, i.e., a bilinear curve is used to describe the elasticity and plasticity of the aluminum with an elastic limit of $\sigma_{0.2}$ and a triple-linear curve for the steel with an elastic limit of $\sigma_{0.01}$. The properties of the stainless steel are adopted from papers of Talja \& Salmi~\cite{1995-Asko-design} and Rasmussen~\cite{2003-Kim-fullrange}, the elastic-plastic properties of aluminum are adopted from the papers of Zha \& Moan~\cite{2003-Yu-experi} and Yun et al.~\cite{ 2021-Yun-fullrange}.

\begin{table}[h]
\begin{center}
\begin{minipage}{0.48\textwidth}
\caption{Elasticity and plasticity properties of the stainless steel and aluminum alloy.}\label{tab1}
\begin{tabular}{@{}lcc@{}} %
\toprule
Properties & Stainless steel & Aluminum\\
\midrule
Density ($\rm{Kg/m^3}$)    & 7750  &  2770  \\
Young's Modulus (GPa)    & 193  & 71  \\
Poisson's Ratio   & 0.31  & 0.33 \\
$\sigma_{0.2}$ (MPa)  & 286  & 300 \\
$\sigma_{u}$ (MPa)  & 627  & 330 \\
$\varepsilon_{u}$   & 0.65  & 0.08 \\
\bottomrule
\end{tabular}
\end{minipage}
\end{center}
\end{table}

\begin{figure}[t]
    \centering
    \includegraphics[width=0.48\textwidth]{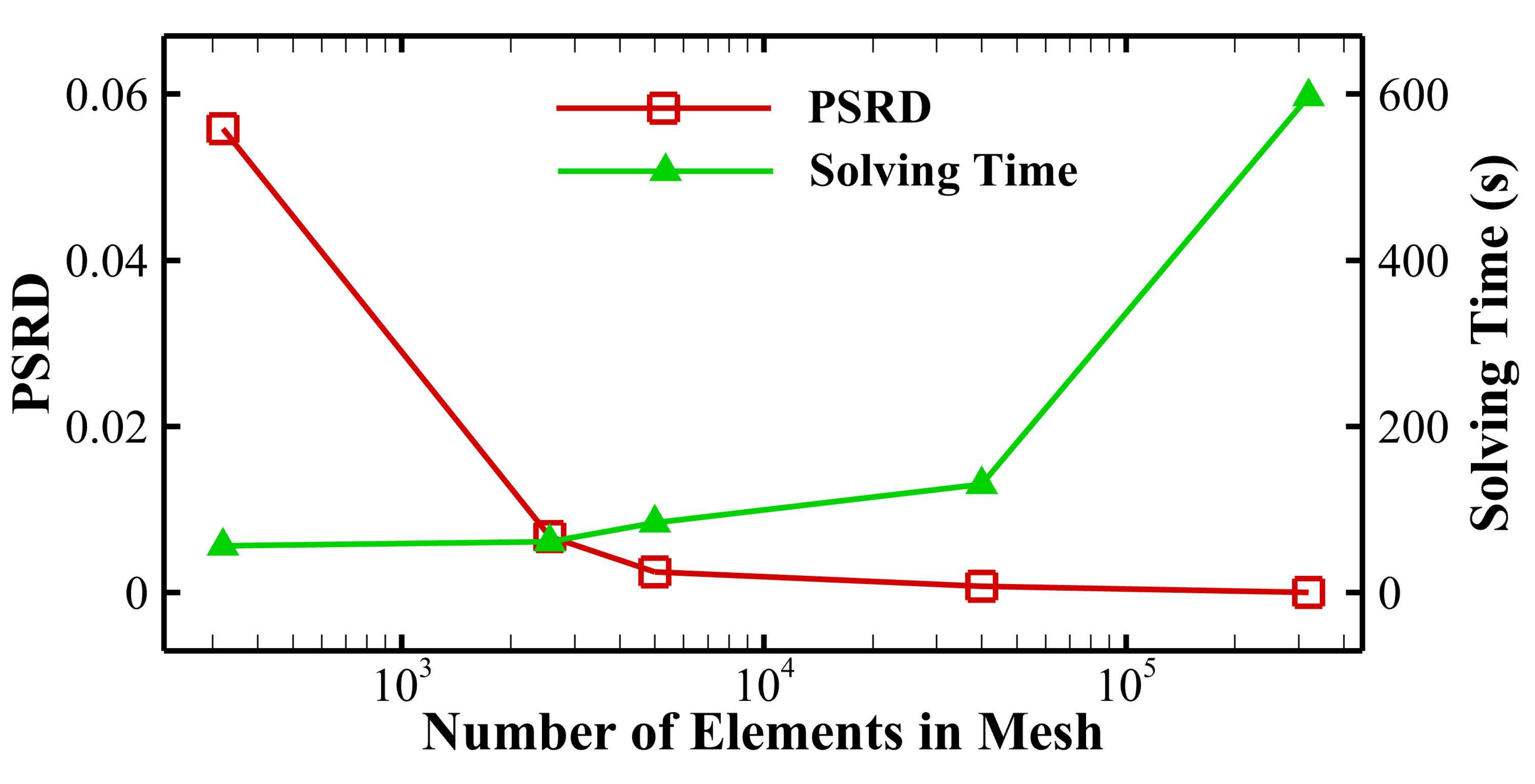}
    \caption{\textbf{Convergence vs solving time.} It shows the PSRD and solving time changes with number of elements of $320, 2560, 5000, 40\,000$ and $320\,000$. With the increase of elements in mesh, the PSRD decreases and the solving time increases.}
    \label{fig3}
\end{figure}

The system is discretized using the finite element method. The mechanical equations are numerically solved by a commercial finite element software ABAQUS to analyze the time-evolution of the reaction force at the fixed end and the stress distribution in the IPC beam under different dynamic strain loadings. Results from the mesh convergence study are presented in Fig.~\ref{fig3}. Given a dynamic strain loading, we monitor the transient stress distributions in the IPC beam for different numbers of elements. Fig.~\ref{fig3} shows the dependence of peak stress relative difference (PSRD) and computational cost on the number of elements, where we take the result obtained by the finest mesh of $320\,000$ as the reference solution. When the system is discretized by more than $40\,000$ elements, the computational cost is greatly increased with little decrease in PSRD. We note that the result of $5\,000$ elements can achieve an accuracy of $99.9\%$ at a much lower computational cost compared to $320\,000$ elements. Therefore, we use $5\,000$ 8-node quadrilateral elements to perform all simulations of this IPC beam under various dynamic loadings to generate data sets for DNO training.

To ensure a good representation of an arbitrarily dynamic strain loading that is imposed on the free end of the IPC beam, the time function of strain is created by drawing from a random Gaussian process. By performing FEA simulations for different inputs of dynamic loading, we can obtain the transient mechanical responses of the IPC beam in terms of the reaction force at the fixed end $F(t)$ and the stress distribution $\sigma(\mathbf{x},t)$ with $\mathbf{x}$ being any location in the IPC beam. These transient outputs of mechanical response to dynamic loading are then combined with the corresponding Gaussian process inputs to form the training data for the DNO model.

\subsection{DNN and DNO}\label{sec2.2}
A simple neural network like a feedforward neural network (FNN) represents the final output by using the nonlinear and linear transform of its original neural inputs. A FNN with $L$ layers can be expressed as,
\begin{equation}
F(x)=G^{(L)}(\cdots (\sigma G^{(3)}(\sigma G^{(2)}(\sigma G^{(1)}(x))))),\label{eq1}
\end{equation}
where $G^{(*)}(x) = W^{(*)}x+b^{(*)}$ and $\sigma$ is the activation function. There are three widely used activation functions including rectified linear activation (ReLU), logistic (Sigmoid), and hyperbolic tangent (Tanh), where Tanh is adopted as the activation function in the DNO model for IPCs.

\begin{figure}[t]
    \centering
    \includegraphics[width=0.48\textwidth]{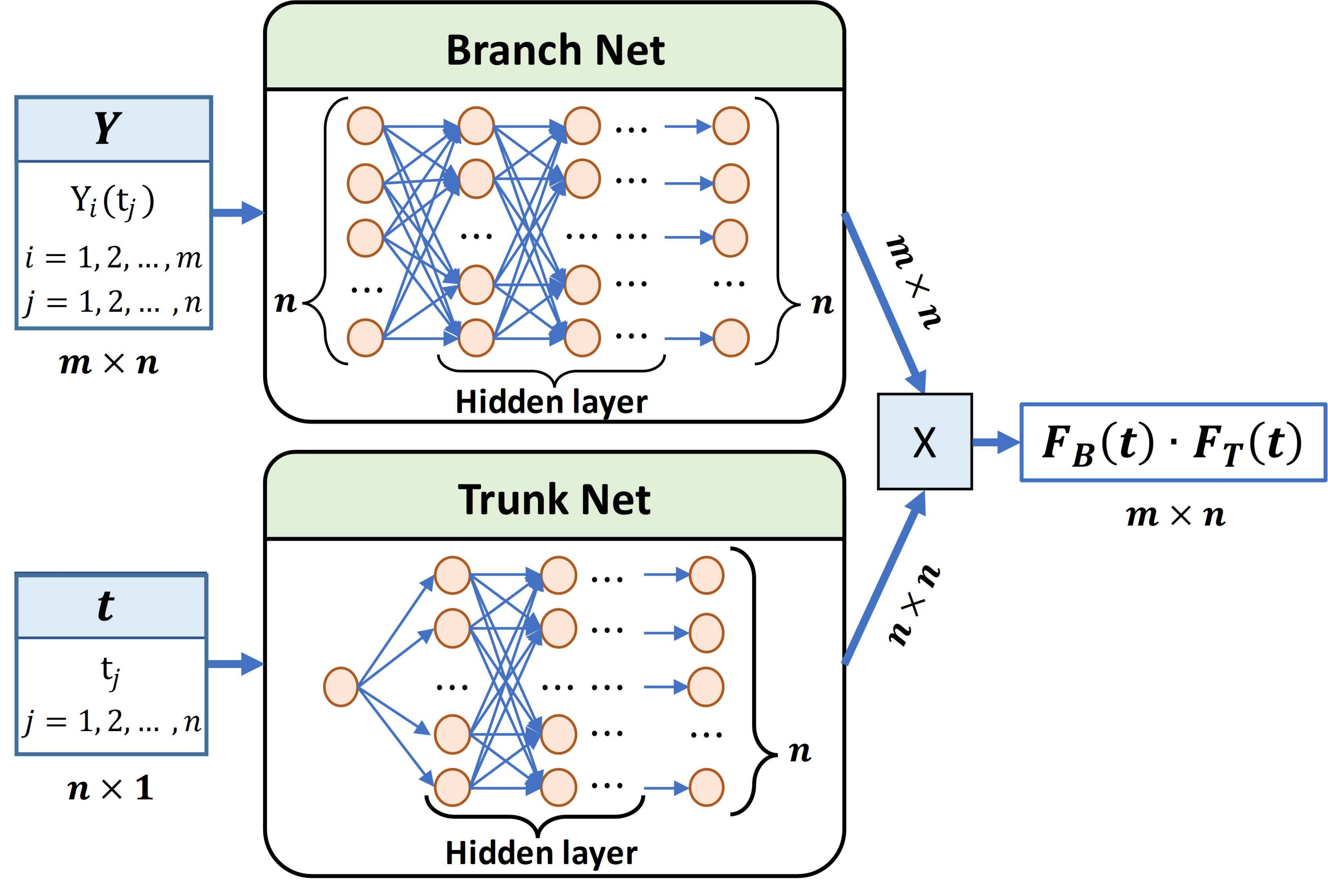}
    \caption{\textbf{Schematic of fast DNO.} The input of data, time series $t$ and time-dependent sequence $Y$, go through two neural networks, trunk net and branch net, respectively. The matrix product of the two neural networks outputs becomes the final output of this deep learning framework.}
    \label{fig4}
\end{figure}

DNO aims to find the mapping relationship between two continuous functions. The schematic of the DNO framework is shown in Fig.~\ref{fig4} as a variant of DeepONet with a trunk net and a branch net in terms of DNNs. To speed up the training process, in our model, the input of trunk net is the unchanged time sequence, vector $t = (t_{1}, t_{2}, t_{3}, ..., t_{N})$, where $n$ stands for discrete time points. The inputs of branch net are the corresponding sequences $Y$ related to this time sequence $t$. It can be expressed as $Y_{i} = (Y_{i}(t_{1}), Y_{i}(t_{2}), Y_{i}(t_{3}), ..., Y_{i}(t_{N}))$, where $i\in[1, m]$ means the $i$-th training dataset and $m$ represents the total number of training datasets.
The entire sequence $Y_{i}$ is fed to the model at a time. In this way, instead of one-by-one training, the training process is sequence by sequence, so that it can shorten the training time effectively. And the setting of the trunk net ensures the continuity of the data and makes it easier to display the time and space dependent features, and makes the mapping relationship more accurate.

The trunk net and the branch net of the framework are both $L$-layer FNN. The trunk net has one neuron in the input layer and $n$ neurons in the output layer. The output of trunk net is a n$\times$1 matrix which can be expressed as,
\begin{equation}
F_{T}(t)=F(t).\label{eq2}
\end{equation}
The branch net has $n$ neurons in the input layer and $n$ neurons in the output layer. The output of branch net is a $m\times~n$ matrix and can be expressed as,
\begin{equation}
F_{B}(t)=F(Y_{i}(t)), ~~~~~{\rm for}~ i=1,2,...,m.\label{eq3}
\end{equation}
The final output of the DeepONet is $U(t) = F_{B}(t)\cdot F_{T}(t)$ with dimension of $m\times~n$.

\subsection{DNO with incremental learning }\label{sec2.3}
Incremental learning is a method of dealing with large-scale data or the gradual accumulation of data. By extracting useful information from newly added data~\cite{2016-Loyala-smart}, it can update the model by modifying the hyperparameters without storing historical data. Incremental learning can be divided into three categories according to its learning tasks: sample incremental learning, feature incremental learning, and category incremental learning. The sample incremental learning method is used in this paper to study the stress distribution of arbitrary cross-sections of the composites.

To improve the accuracy of prediction, especially in the case of limited data, incremental learning shows great advantages. Although the training data in this paper comes from FEA, increasing the training set also requires more computational cost. Since the stress distribution on the same cross-section has a certain continuity and follows the same physical laws, the data of different nodes can be used as incremental data for model training.

\begin{figure}[t]
    \centering
    \includegraphics[width=0.48\textwidth]{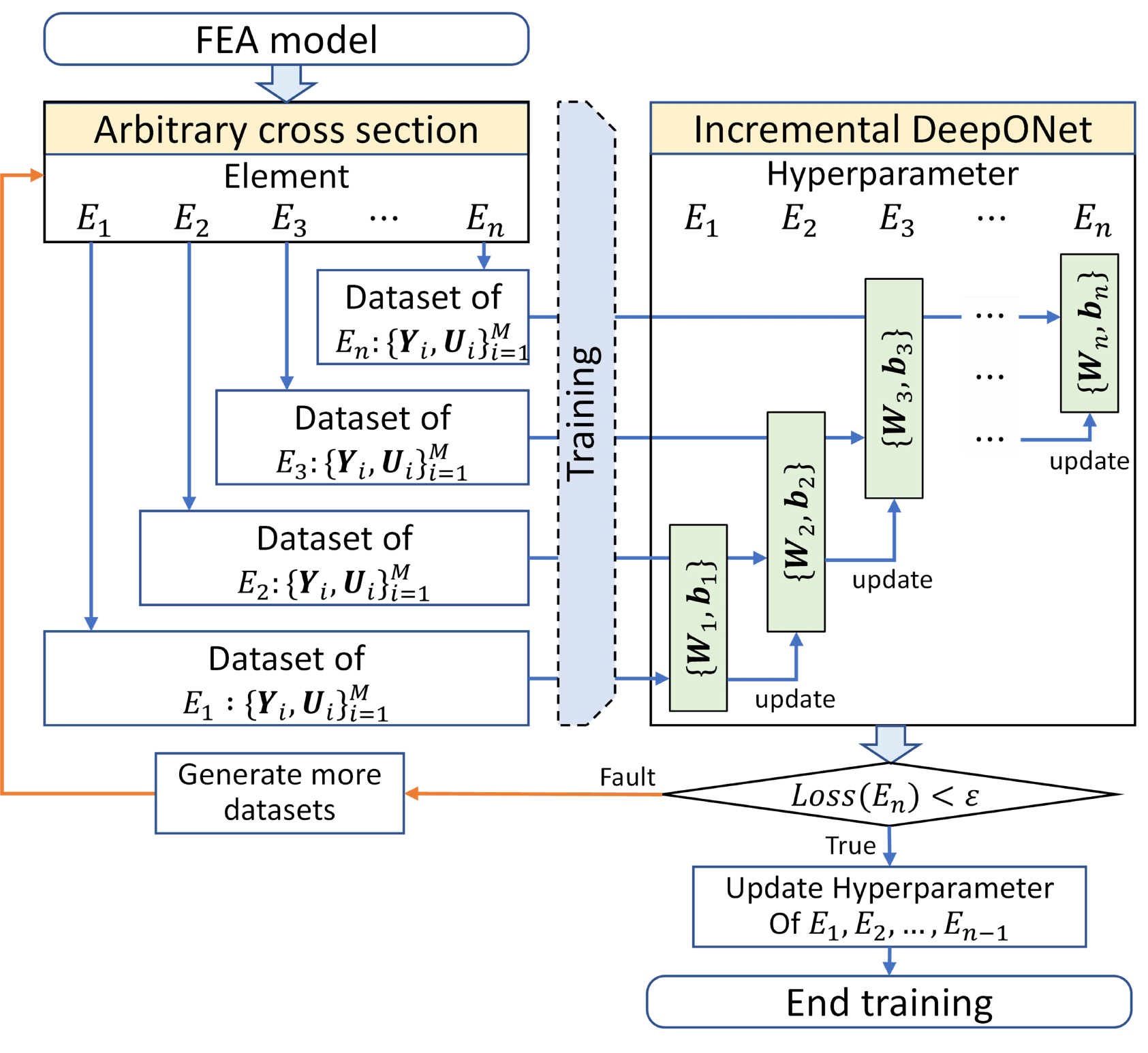}
    \caption{\textbf{Float of incremental learning in DeepONet.} For any certain cross-section of the structure, it is described by $n$ elements in the FEA model. $n$ can be different due to different directions of different cross-sections. At the beginning, $m$ sets of data for each element were obtained from FEA. The training starts at element 1 with $m$ sets of data, the element 2 is trained based on the hyperparameters for element 1. So on and so forth, $n$th element is trained based on the hyperparameters for $(n-1)$th element, approximately the same as it is trained with $n\cdot m$ data.}
    \label{fig5}
\end{figure}

The flowchart of incremental learning is shown in Fig.~\ref{fig5}. At the beginning of training, $m$ datasets are collected by the FEA computation for each element within a cross-section of the composites. After the DNO for the first element is trained by the M datasets. The second element can be trained based on the previous hyperparameters of the first element. And the hyperparameters are kept updated until the DeepONet model for the last ($n$th) element is trained. If the accuracy reaches the requirement, the system goes back to the first element model to renew all hyperparameters until the $(n-1)$th element. By incremental learning, The entire DeepONet model is equivalent to being trained with $n\cdot m$ datasets. If the final loss still fails to meet the requirement, then extra data needs to be generated by the FEA model.

\section{Results}\label{sec3}
\begin{figure}[t]
    \centering
    \includegraphics[width=0.49\textwidth]{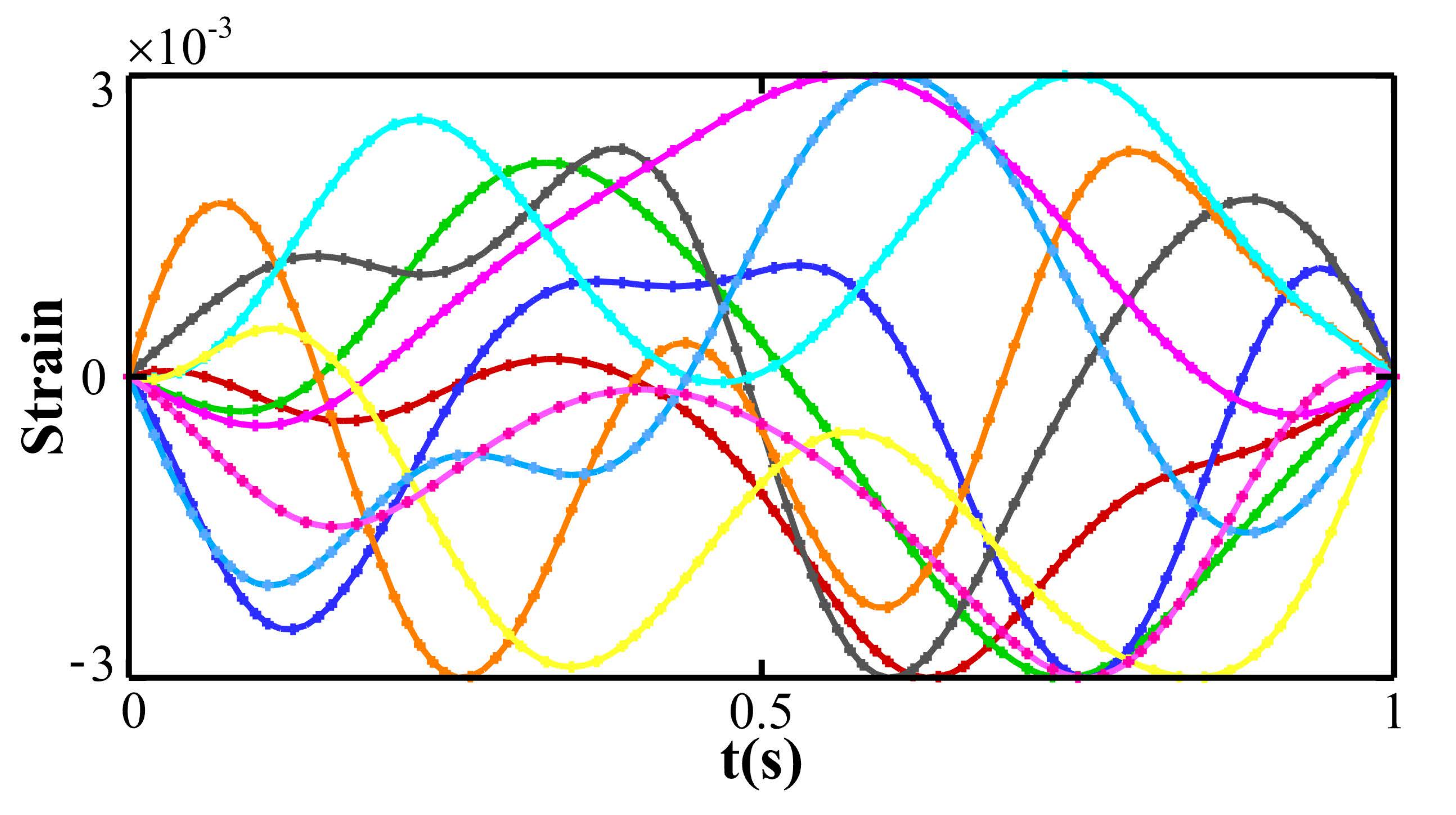}
    \caption{\textbf{Typical dynamic strain loading}. Arbitrary dynamic strain loading is created by drawing a smooth time-dependent function between 0 to 1 seconds from a Gaussian random process.}
    \label{fig-epsilon}
\end{figure}

The arbitrarily dynamic strain loading imposed at the free end of the IPC beam is set to be a randomly time-varying function drawn from a Gaussian random process. We define the dynamic strain loading as $\epsilon(t)\sim GP(\mu,\sigma\cdot k(t, t'))$, where the mean of the strain $\mu = 0$ and the standard deviation $\sigma = 3\times10^{-3}$. An exponential kernel function in the form of $k(t,t') = \exp(-20\|t-t'\|^2)\cdot \sin(\pi t)$ is used to ensure a smooth strain loading process, in which the term $\sin(\pi t)$ sets that the stain loading always starts from a zero strain at $t = 0~{\rm s}$ and ends with a zero strain at $t = 1~{\rm s}$.
Fig.~\ref{fig-epsilon} displays the typical dynamic strain loading created by the Gaussian random process.
The mechanical responses of our interest are the time evolution of the total force $F(t)$ generated at the fixed end and the normal stress along the $x$-direction $\sigma_{xx}(\mathbf{x},t)$ inside the IPC beam, where $\mathbf{x}$ denotes an arbitrary location in the IPC beam.

We perform $5\,000$ FEA simulations for the IPC beam subject to $5\,000$ different dynamic strain loadings to collect enough data to train the DNO model. Each simulation runs from $t=0~{\rm s}$ to $t = 1~{\rm s}$ with a sampling interval $\Delta t = 0.01~{\rm s}$, which generates one data set consisting a time sequence $\{t_j\}$ with $j=1,2,\cdots,101$, a time-dependent input $\{Y_i(t_j)\}=\epsilon(t_j)$, and a corresponding time-dependent output $\{\mathbf{U}(t_j)\}=\{F(t_j),\sigma(\mathbf{x},t_j)\}$ with $\mathbf{x}$ being an arbitrary location inside the IPC beam. Both the branch net and the trunk net of DNO have one input layer, five hidden layers, and one output layer, with an architecture of $[101]\cdot[200]\cdot[200]\cdot[300]\cdot[200]\cdot[200]\cdot[101]$, in which the integers represent the number of neurons in those layers. The loss function is defined as the mean squared error between the true value in the FEA data set and the network prediction. Adam optimizer with a small learning rate $0.001$ is applied to train the DNO model using $4\,900$ data sets until the desired accuracy is achieved or a predefined maximum iteration of $2\,500\,000$ is reached.

\begin{figure}[t]
    \centering
    \includegraphics[width=0.49\textwidth]{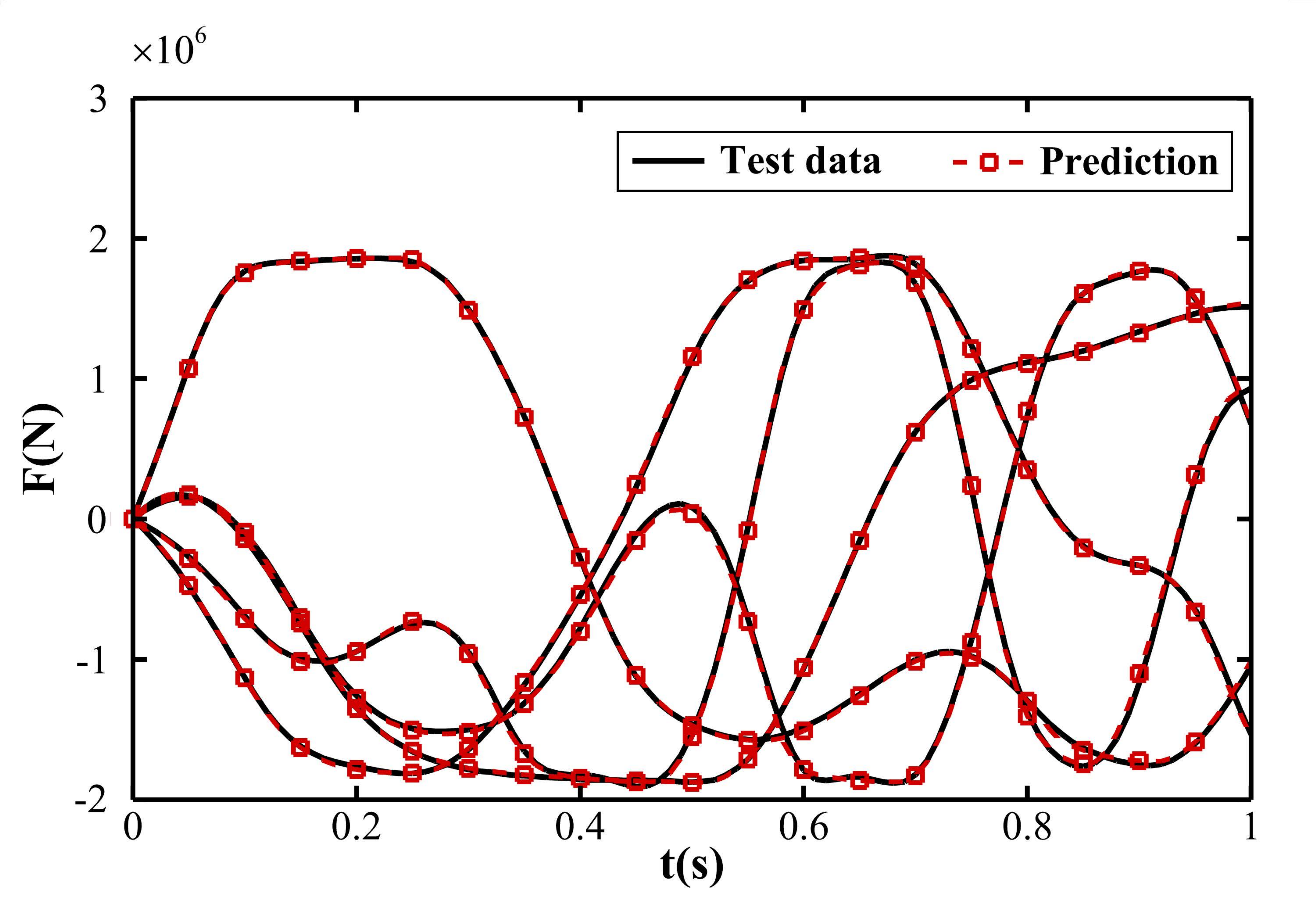}
    \caption{\textbf{Prediction of transient force response}. The time-evolution of reaction force $F(t)$ in five typical time-dependent sequences from 0 to 1 seconds are plotted. Test data are FEA results, while prediction is made by the trained DNO model. }
    \label{fig6}
\end{figure}
\subsection{Prediction of force response}\label{sec3.1}
We first investigate the performance of the DNO model on predicting the time evolution of reaction force $F(t)$ at the fixed end of the IPC beam under different dynamic strain loading. We use $4\,900$ FEA data sets to train the DNO model and use the rest $100$ data sets to test the accuracy of the trained DNO model. The accuracy is defined based on a relative error using $L_2$ norm,
\begin{equation}\label{eq:acc}
  \chi = \left( 1 - \frac{\lVert F-F_{\rm ref}\rVert_2}{\lVert F_{\rm ref}\lVert_2}\right)\times 100\%,
\end{equation}
where $F$ represents the prediction and $F_{\rm ref}$ the reference values. Fig.~\ref{fig6} shows a comparison of $F(t)$ between FEA results and DNO prediction, where we compute the relative mean square error to quantify the prediction accuracy. The result shows that the average prediction accuracy of the DNO model on the $100$ output sequences is $\chi=98.04\%$, and the best prediction accuracy can reach $\chi=99.58\%$ in the test data.

\begin{figure}[t]
    \centering
    \includegraphics[width=0.49\textwidth]{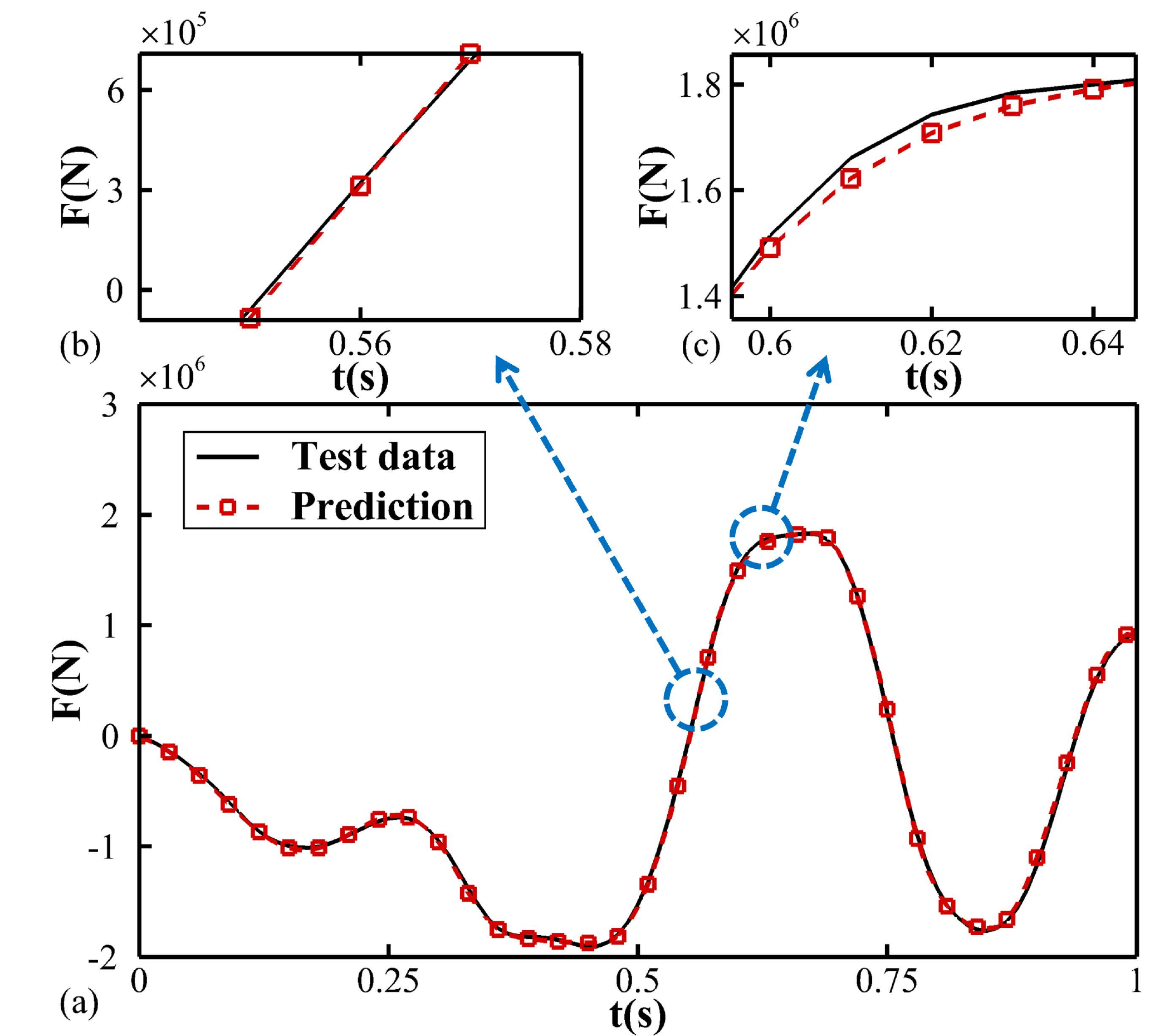}
    \caption{\textbf{Distribution of prediction errors.} (a) Comparison of transient force response to a dynamic strain loading between DNO prediction and FEA simulation. (b) Zoom-in view for elastic deformation stage. (c) Zoom-in view for plastic deformation stage.}
    \label{fig7}
\end{figure}

Because the IPC beam may have nonlinear plastic deformation under large strains, the prediction error is not a uniform distribution over time when a dynamic strain loading is imposed. In general, given a dynamic strain loading, the composites structure experiences both elastic and plastic deformation. We select one specific case to show where the DNO model may have large prediction errors. Fig.~\ref{fig7}(a) plots the predicted $F(t)$ compared with the corresponding FEA result, with a zoom-in view in Fig.~\ref{fig7}(b) for the stage of linear elastic deformation and Fig.~\ref{fig7}(c) for the stage of nonlinear plastic deformation. We observe small prediction errors in Fig.~\ref{fig7}(b) and large prediction errors in Fig.~\ref{fig7}(c). The reason could be that some subdomains of the composite beam are in a plastic tension or compression stage while other subdomains are in an elastic tension or compression stage, which creates a very complicated nonlinear mechanical response that is harder to learn in the training data.

\subsection{Prediction of cross-section stress distribution}\label{sec3.2}
\begin{figure}[t]
    \centering
    \includegraphics[width=0.49\textwidth]{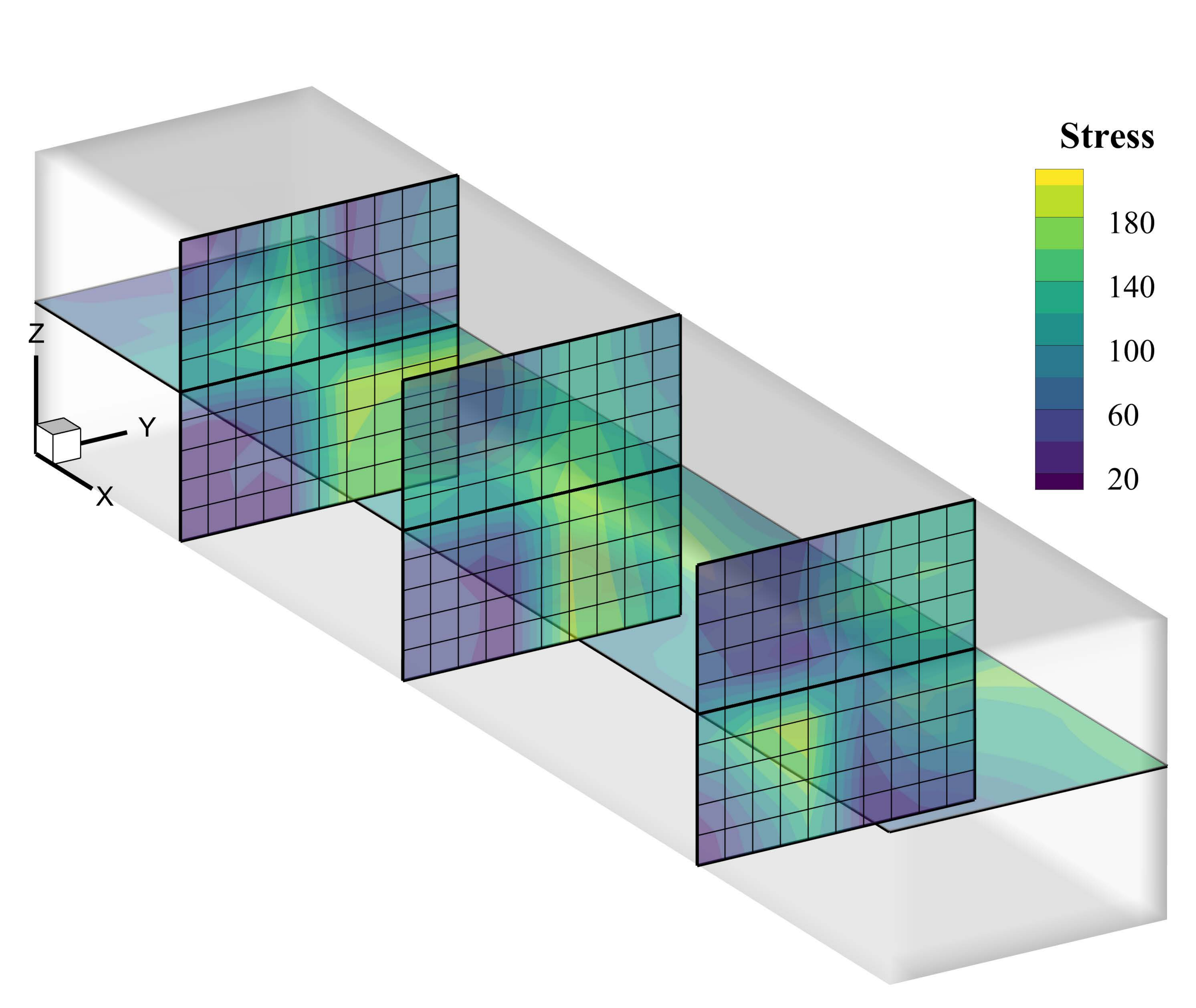}
    \caption{\textbf{Cross-section stress distribution of the composites.} The figure shows the stress distribution on a random transverse section and three longitudinal sections of the composites structure at $t = 0.86~{\rm s}$ under a dynamic strain loading. Stress shown here refers to the normal stress along the x-direction $\sigma_{xx}$.}
    \label{fig8}
\end{figure}

The relationship between stress and strain is an important characteristic of a material. For traditional materials, it reveals many properties such as Young's modulus, yield strength, and ultimate tensile strength. It can also reflect the mechanical properties of composite materials.
Using the same method, the stress distribution of different cross-sections can be predicted by the trained DNO model. By training the DNO model based on stress distribution data on selected cross-sections, the DNO model can be used to predict the time-evolution of stress distribution at any selected cross-sections as shown in Fig.~\ref{fig8}.

The DNO model is trained using the incremental learning algorithm as described in Section~\ref{sec2.3}.
On a certain selected cross-section of the composite beam, although the two constituent materials differ in elasticity and plasticity, they both obey the same solid mechanical equations. In this case, the data of one element within this cross-section can be treated as supplementary data from another element. Therefore, incremental learning can be introduced in the training process of stress distribution. It means that after training the first element, the model does not need to restart from a randomly initialized network to train the next element. Instead, the DNO model can keep the previous hyper-parameters of each neuron, and train the next element with the data accompanying it.

\begin{figure*}[h!]
    \centering
    \includegraphics[width=0.7\textwidth]{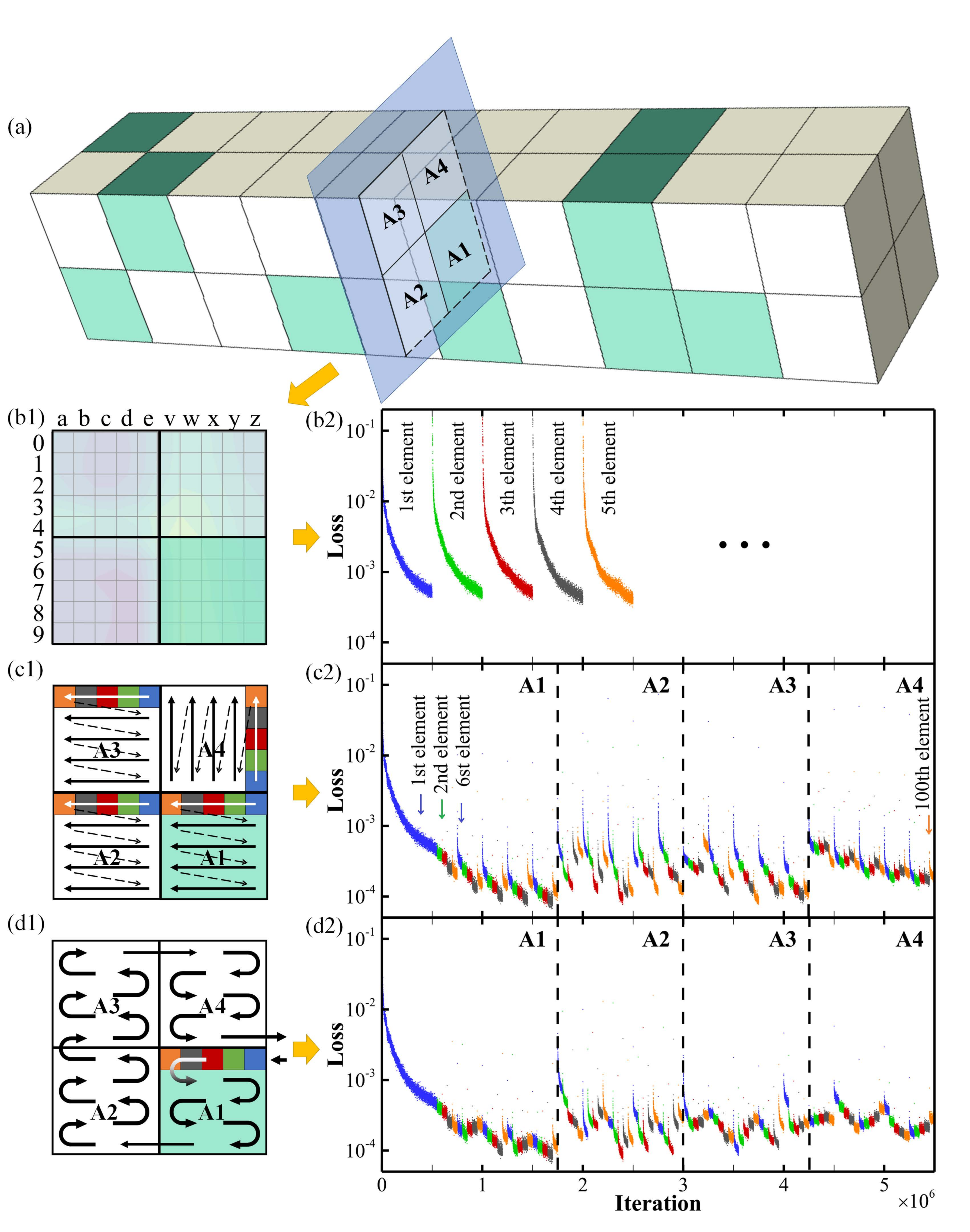}
    \caption{\textbf{Comparison of the training loss with the number of iterations without incremental learning and with different incremental learning methods.}
    (a) The selected cross-section and its material arrangement. (b1) The selected cross-section and its elements naming table. (b2) The training loss with the number of iterations without incremental learning. (c1) Incremental learning with the default element order. (c2) The training loss with the number of iterations with incremental learning under default training order. (d1) Incremental learning with the rearranged element order. (d2) The training loss with the number of iterations with incremental learning under rearranged training order.
    }
    \label{fig9}
\end{figure*}

Fig.~\ref{fig9}(a) presents a cross-section selected for analysis, comprising 100 elements in the FEA model. The cross-section is composed of four areas that are labeled as $A1$, $A2$, $A3$ and $A4$, where $A1$ with green color represents aluminum alloy and $A2$, $A3$, and $A4$ with white color are made of stainless steel. To facilitate identification, a naming table for each element is provided in Fig.~\ref{fig9}(b1). The training process for this cross-section is performed in two ways: 1) a regular training algorithm where each element is trained independently, and 2) an incremental learning (transfer learning) where the network of next element uses the network variables of previous elements as initial condition. Fig.~\ref{fig9}(b2) shows the training loss for the regular training algorithm as a function the number of iterations, where each element is trained over $500\,000$ iterations to reach a training loss of $5\times10^{-4}$. Thus, the entire training process requires $50\,000\,000$ iterations in total, which takes approximately $84$ hours using a PC with an Intel i9-10900K CPU. However, the incremental learning algorithm can optimize the training process by updating the hyperparameters of element based on the results of the previously trained element.

Each element in the FEA model is identified by a sequence number, and the elements in the selected cross-section are arranged in a specific order as depicted in Fig.~\ref{fig9}(c1). The route starts from $z5$, the first element in $A1$, and moves sequentially to $y5$, $x5$, $w5$, and $v5$. Subsequently, it moves back to $z4$ and follows the same pattern until it reaches the last element in $A1$, which is $v9$. The route then proceeds to the $A2$ area, starting from $e5$ and finishing at $a9$. The route for $A3$ starts at $a4$ and ends at $e0$, and the route for $A4$ follows a different path, starting from $z4$ and moving to $z3$, $z2$, $z1$ and $z0$. It then moves back to $y4$, $y3$, $y2$, $y1$, $y0$, and finally ends at $v0$. This route is called the default route. Fig.~\ref{fig9}(c2) illustrates the training loss for each element when following the default route.

With an incremental learning process, the DNO model begins training by taking $500\,000$ iterations to reach a training loss of $5\times10^{-4}$ for the first element. The second element requires fewer iterations, reaching a smaller training loss with only $50\,000$ iterations. To demonstrate the process of implementing the incremental learning, all elements (except the first) are trained for $50\,000$ iterations, based on the result of the previously trained element. In Fig.~\ref{fig9}(c2), a small gap in training loss is observed between each set of five element in $A1$, while a larger gap is observed between the last element in the $A1$ area (v9) and the first element in the $A2$ area (e5). These observations confirm that the chosen cross-section follows the same physical laws and that the DNO model with an incremental learning can successfully learn these laws. However, different areas ($A1$, $A2$, $A3$ and $A4$) have different material properties, resulting in different mechanical states within them. The similarity between elements is better for neighboring elements in the same area, leading to gaps between every five elements in the loss curve when following the default route. If the element can be trained based on the result of an element with a better similarity, the process can be faster and the results can be more accurate. The loss curve in Fig.~\ref{fig9}(d1) shows more continuity and smoother training loss changes than Fig.~\ref{fig9}(c1) due to the rearranged route for incremental learning. Removing the iteration limit and training until each element has a loss less than $2\times10^{-4}$ shows that the rearranged route is $7\%$ faster than the default route while maintaining the same accuracy.

\begin{figure*}[h!]
    \centering
    \includegraphics[width=0.98\textwidth]{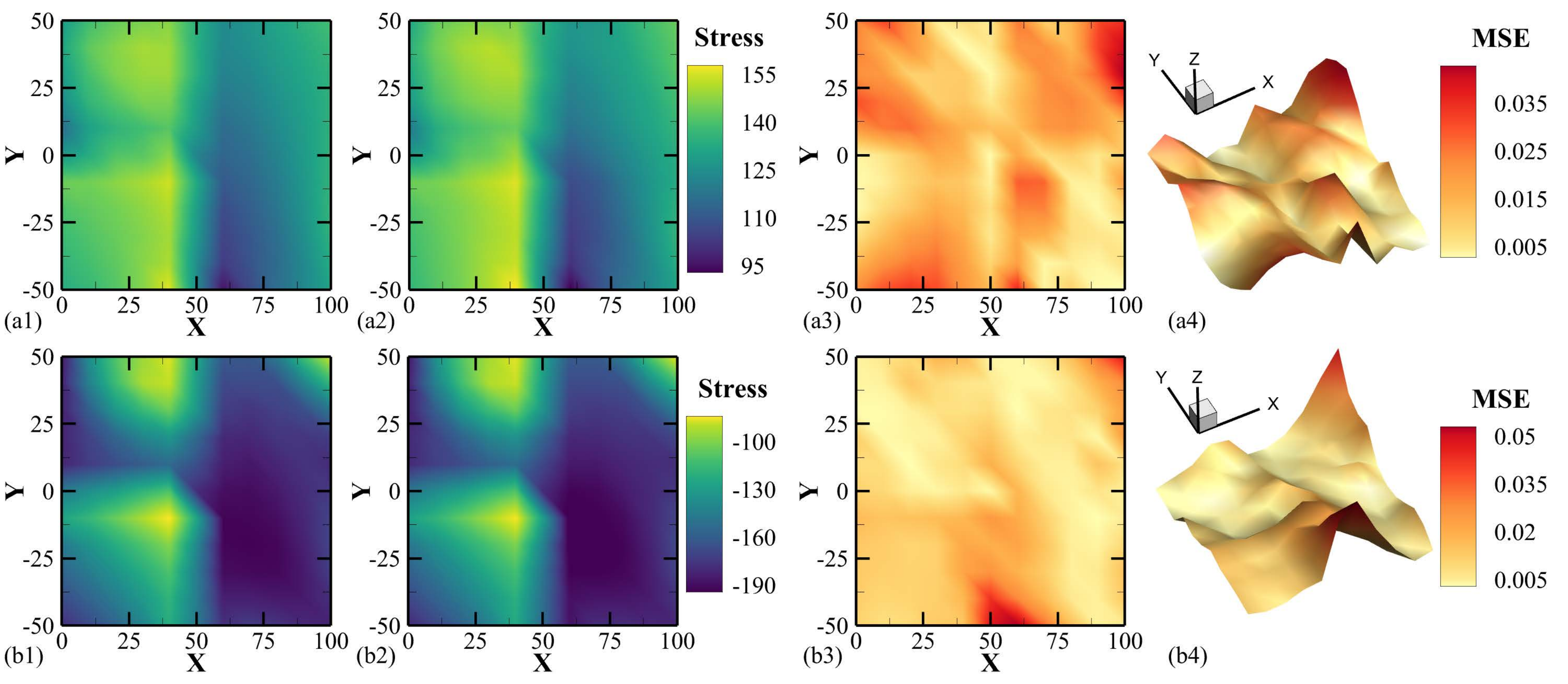}
    \caption{\textbf{Prediction of cross-section stress distribution and error distribution.} (a1)\&(b1) shows the stress distribution of selected cross-section from FEA model at 0.5 s and 1.0 s respectively. (a2)\&(b2) illustrates the prediction of stress distribution of selected cross-section from DNO model at 0.5 s and 1.0 s respectively. (a3),(b3) shows the 2D prediction error distribution within cross-section at 0.5 s and 1.0 s respectively. (a4),(b4) shows the 3D prediction error distribution within cross-section at 0.5 s and 1.0 s respectively.}
    \label{fig10}
\end{figure*}

Figs.~\ref{fig10}(a1)\&(a2) show the stress distribution at the center cross-section (perpendicular to the $x$-axis) of the IPC beam at $t = 0.5~{\rm s}$ for the test data and the DNO predicted result, respectively. The average predicted accuracy of the DNO model is $\chi=98.18\%$. Fig.~\ref{fig10}(a3) shows the predicted error distribution within the cross-section, which is relatively smaller in the central region and increases towards the edge of the structure. As the structure has greater uniformity in the central area and more heterogeneity near the edge, the highest prediction error occurs in the top-right corner as shown in Fig.~\ref{fig10}(a4). Additionally, the presence of a fixed end as the only boundary constrain gives enough freedom to the rest of the structure, resulting in the structure bending and twisting under compression, adding complexity to the structure dynamics and leading to unique error distributions on the right part of the beam.

The dynamic strain loading curve, as shown in Fig.~\ref{fig6}, is designed to have zero strain imposed on the free end of the beam at the end of loading $t=1~{\rm s}$. Although the entire structure has returned to its original length, residual stresses remain in the structure because of plastic hardening. This is evident from Fig.~\ref{fig10}(b1), which shows the distribution of residual stress in the composite structure at $t = 1~{\rm s}$. During dynamic loading tests, the imposed strain may cross the zero point multiple times, leading to different stresses in the structure after each return to its original length due to plastic deformation. The time and space dependence of the loading process and corresponding mechanical response of heterogenous materials makes it challenging to predict material behavior using traditional neural networks.
However, the trained DNO model provided accurate predictions of stress distributions on various cross-sections, with accuracy ranging from $\chi=95\%$ to $\chi=99\%$ compared to FEA results. This highlights the ability of the DNO model to serve as a surrogate of FEA models in predicting the behavior of composites subjected to dynamic loading.

\subsection{Training speed}\label{sec3.3}
The traditional DeepONet models available in a widely used DeepXDE library~\cite{2021-Lu-DeepXDE} are implemented in TensorFlow and concentrate on training the data by points. The current DNO model is implemented in JAX and uses sequence-to-sequence training to improve the DNO training efficiency.
JAX is a language for expressing and composing transformations of numerical programs~\cite{2018-James-JAX} and brings high-performance computing capabilities. By fixing the time sequence in the trunk net and improving the input form of the corresponding data in the branch net, the DNO model is trained by sequence-to-sequence rather than by points used in traditional DeepONet models. Together with an incremental learning algorithm, it has an extremely fast training speed.

On the prediction of the total reaction force of the IPC beam, it takes $205$ seconds for training the DNO model, which is nearly 10 times faster than the original DeepONet model at the same average prediction accuracy of $98.17\%$. The DNO model shows higher training efficiency in predicting the stress distribution of the IPC beam. Here, a higher training loss threshold is used to compare the training wall time of different methods. Table~\ref{tab2} compares the wall time for training a traditional DeepONet model, a DNO model without incremental learning (DNO basic), and a DNO model with incremental learning (DNO incre.). All training processes are performed on the same PC with an Intel i9-10900K CPU, and training is set to stop when the training loss is below $5\times10^{-3}$.

\begin{table}[h]
\begin{center}
\begin{minipage}{0.49\textwidth}
\caption{Comparison of computational efficiency for different architectures and training methods, where $M*N$ indicates a DNN with $N$ layers and $M$ neurons for each layer.}\label{tab2}
\begin{tabular}{@{}llccc@{}}
\toprule
& \multicolumn{4}{@{}c@{}}{\hspace{1.2cm}{Wall time (min)}} \\\cmidrule{3-5}
No. & Arch. & DeepONet & DNO (basic) & DNO (incre.)\\
\midrule
1 & 200*3  & 2094.9  & 284.9 & 7.7 \\
2 & 200*4  & 2418.3 & 310.3 & 8.0 \\
3 & 200*5  & 2378.9 & 324.8  & 9.5 \\
4 & 200*6  & 2085.0  & 328.8 & 9.2 \\
5 & 200*7  & 2204.4 & 358.3 & 10.3 \\
6 & 200*8  & 2710.2 & 530.1  & 12.2 \\
\bottomrule
\end{tabular}
\end{minipage}
\end{center}
\end{table}

When neural networks have more neurons and layers, longer training times are required, as demonstrated by Table~\ref{tab2}. Specifically, the DNO model's training wall-time decreases as the number of hidden layers increases, indicating that the neural networks with more layers requires fewer iterations to achieve the same training loss.
Furthermore, the ratio of training wall-time for the DeepONet model built with TensorFlow to that of the DNO model built with JAX (i.e., $r = \rm{T_{DeepONet}/T_{DNO}}$) increases as the number of hidden layers increases. In case No.~1, $r$ is $2094.9/284.9=7.35$, while in case No.~6, $r$ is $2710.2/530.1=5.11$. Compared to TensorFlow implementations, JAX-based models are trained much faster, particularly for complex deep neural networks.

Table~\ref{tab2} also demonstrates that the DNO model with incremental learning achieves a speed-up of approximately $40$ times compared to the DNO model without incremental learning, using the same deep learning framework. Considering that the JAX-based DNO model without incremental learning still has an approximately 5 times speed-up compared to the traditional DeepONet model implemented in TensorFlow, the proposed DNO model with incremental learning, implemented in JAX, could achieve a training speed-up of $200$ times.

\subsection{Model analysis}\label{sec3.4}
The efficiency and prediction accuracy of the DNO model can be affected by the different architectures of the branch and trunk networks, the size of the training dataset, and the number of training iterations. These factors are analyzed in this section to achieve better performance of the DNO model. Also, the robustness and the generalizability of the DNO model will be investigated.

\subsubsection{Architecture of DNO}\label{sec3.4.1}
The architecture of the deep learning model has a crucial impact on the prediction ability. It is in general difficult to find the best architecture. However, we can still test the effect of the different numbers of layers and the different numbers of neurons in each layer on the model to get a better solution. A group of DNO models with different architectures are tested with the same optimizer and $500\,000$ training iterations as shown in Table~\ref{tab3}. For vanilla neural networks, only the cases with less than 10 hidden layers are tested here.

\begin{table}[h]
\begin{center}
\caption{Influence of architecture and batch size}\label{tab3}
\begin{tabular}{@{}cccc@{}}
\toprule
No. & Architecture  & Batch size  & Accuracy\\
\midrule
1 & 200      & 64 & 66.05\% \\
2 & 200*2      & 64 & 97.90\%  \\
3 & 200-300-200      & 64  & 97.38\%  \\
4 & 200*3      & 64  & 98.17\%  \\
5 & 200*4      & 32  & 97.18\%  \\
6 & 200*4      & 64  & 98.12\%  \\
7 & 200*2-300-200*2      & 32  & 97.94\%  \\
8 & 200*2-300-200*2      & 64  & 97.68\%  \\
9 & 200*8      & 64  & 97.31\%  \\
\bottomrule
\end{tabular}
\end{center}
\end{table}

Table~\ref{tab3} shows that too many or too few hidden layers result in less accurate predictions at fixed training iterations. With 3 hidden layers (architecture No.~4), the model performs better than other architectures. The difference in batch size has an impact on the prediction accuracy but has no obvious relationship. However, with a relatively large batch size, the training loss of the model is more stable.

\subsubsection{Size of training dataset}\label{sec3.4.2}
\begin{figure}[ht!]
    \centering
    \includegraphics[width=0.49\textwidth]{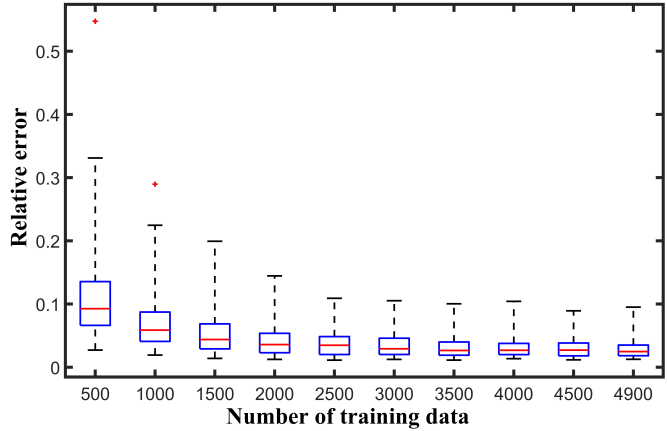}
    \caption{\textbf{Box plot of relative error with different numbers of training data.} On each box, the center mark represents the median of relative error of all test data and predictions, and the bottom and top edges of the box represent the 25th and 75th percentiles of the relative error. The maximum whisker length is specified as 4 times the interquartile range. The `+' marks stand for the extreme results.}
    \label{fig11}
\end{figure}

We evaluate the relationship between the size of the training data set and the accuracy of the DNO model. We note that the improvement of the accuracy by increasing the size of the training data set is not unlimited. Fig.~\ref{fig11} shows how the relative error of the trained DNO model changes with the number of training data.

We observe from Fig.~\ref{fig11} that the mean relative error changes from $9.26\%$ to $2.48\%$ when the number of training data is increased from $500$ to $4\,900$. The prediction accuracy of the trained DNO model on testing data sets is not stable when only $500$ training data sets are used. Although the mean relative error is $9.26\%$, the upper bound reaches $33.11\%$ associated with the worst error of $54.73\%$. As the number of training data set exceeds $3\,000$, the training error remains around $2.4\%$ to $2.6\%$, which suggests that two to three thousand training data sets could be sufficient enough to obtain an accurate DNO model in this problem.

\subsubsection{Time-extended prediction}\label{sec3.4.3}
As mentioned in Section~\ref{sec3.3}, the output of the nonlinear composite system is highly time and space dependent, making it hard to predict the time-extended output. However, the DNO model aims to approximate the hidden mapping operator. If the DNO model is well-trained, it should show a certain time-extended ability.

\begin{figure}[t]
    \centering
    \includegraphics[width=0.49\textwidth]{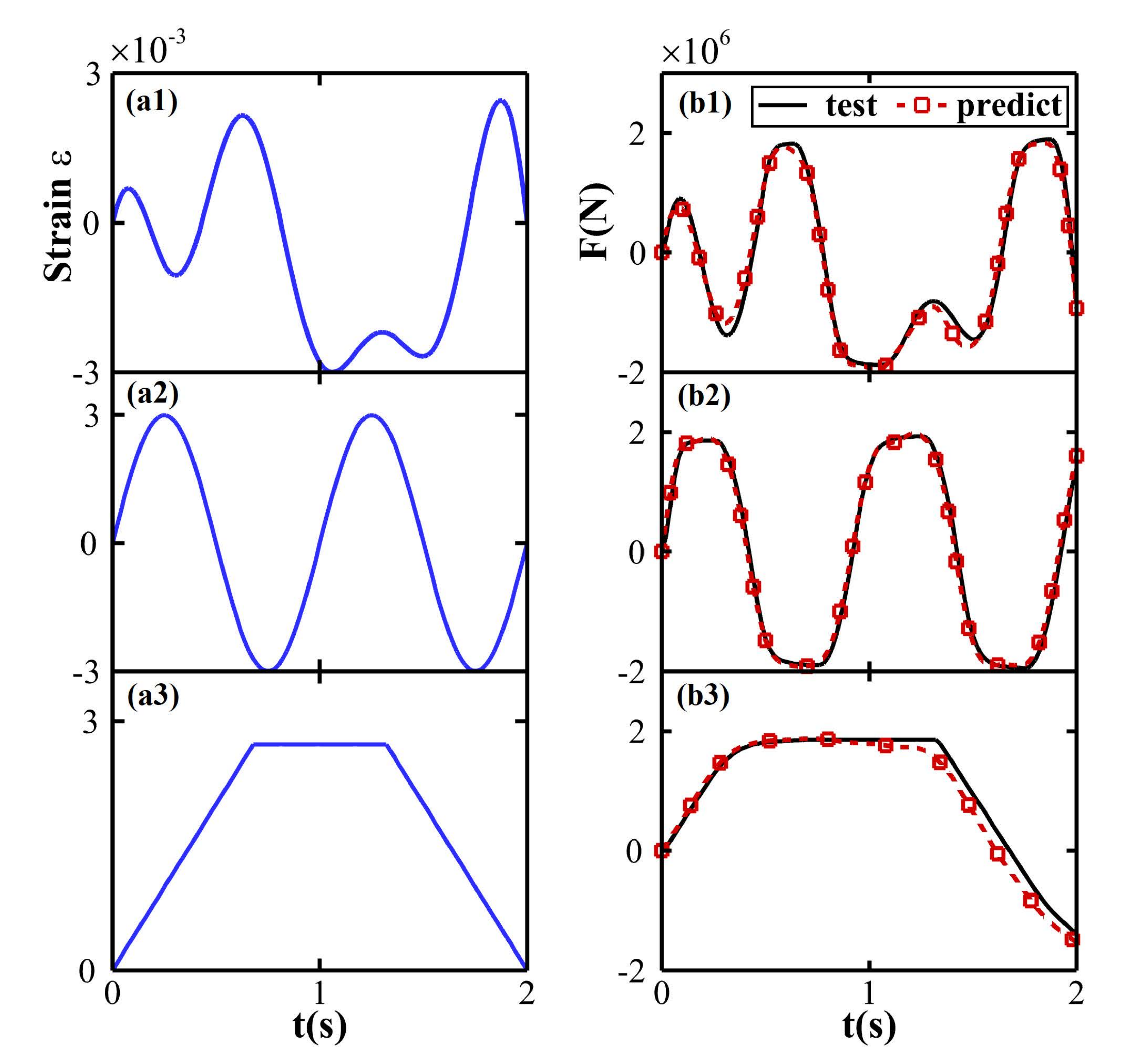}
    \caption{\textbf{Time-extended prediction.} (a) Inputs of dynamic strain loading and (b) transient reaction force at the fixed end, where (a1)\&(b1) are the results of a sequence of random Gaussian process, (a2)\&(b2) are the results of a sinusoidal strain loading, and (a3)\&(b3) are the results of a sequence of piecewise linear strain loading.}
    \label{fig12}
\end{figure}

The proposed DNO model is trained based on a $1~{\rm s}$ interval using random Gaussian process inputs and corresponding outputs. To test the DNO model on time-extended prediction, three cases of 2 seconds dynamic strain loading are tested. Case 1 is a sequence of random Gaussian processes, case 2 is a sinusoidal strain loading with a peak amplitude of $3\times10^{-3}$, and case 3 is a sequence of piecewise linear strain loading, as displayed in Fig.~\ref{fig12}.

The sequence of case 1 is drawn from a similar random Gaussian process of 2 seconds as shown in Fig.~\ref{fig12}(a1), which shares the same kernel function as the inputs of training data. Fig.~\ref{fig12}(b1) shows the prediction of the corresponding force response with an accuracy of $\chi=89.56\%$, indicating that the DNO model can be used to predict time-extended data at a reasonable well accuracy. Case 2 and case 3 are created to verify the prediction ability of the DNO model for an arbitrary input that is not included in the training data set. Fig.~\ref{fig12}(a2) shows the two-period sinusoidal input with a peak strain of $3\times10^{-3}$. In this case, the structure undergoes elastic and plastic deformation successively. It can be observed from Fig.~\ref{fig12}(b2) that the force response of the second period is different from that of the first period, which is due to the residual stress in the composite beam after plastic deformations. The DNO model captures the key details and provides the prediction of $F(t)$ with an accuracy of $\chi=92.81\%$. Fig.~\ref{fig12}(a3) displays an example of piecewise linear strain loading, with corresponding force response $F(t)$ shown in Fig.~\ref{fig12}(b3).

The test results of the three different types of time-extended dynamic loading suggest that the DNO model can predict the transient mechanical responses of the IPC beam subject to different types of dynamic loading. By training the DNO model with one second of data, the trained DNO model can be used to predict longer-time mechanical responses at a reasonable good accuracy.
However, this time-extended ability has its limitation. Since the length of the training sequence is fixed and finite, if the input time-extended sequence is much longer than the training sequence, then at the same length, the input can no longer be well described, especially when faced with an input that has a relatively higher frequency of amplitude changes.

\subsubsection{Robustness of DNO}\label{sec3.4.4}
\begin{figure}[ht!]
    \centering
    \includegraphics[width=0.48\textwidth]{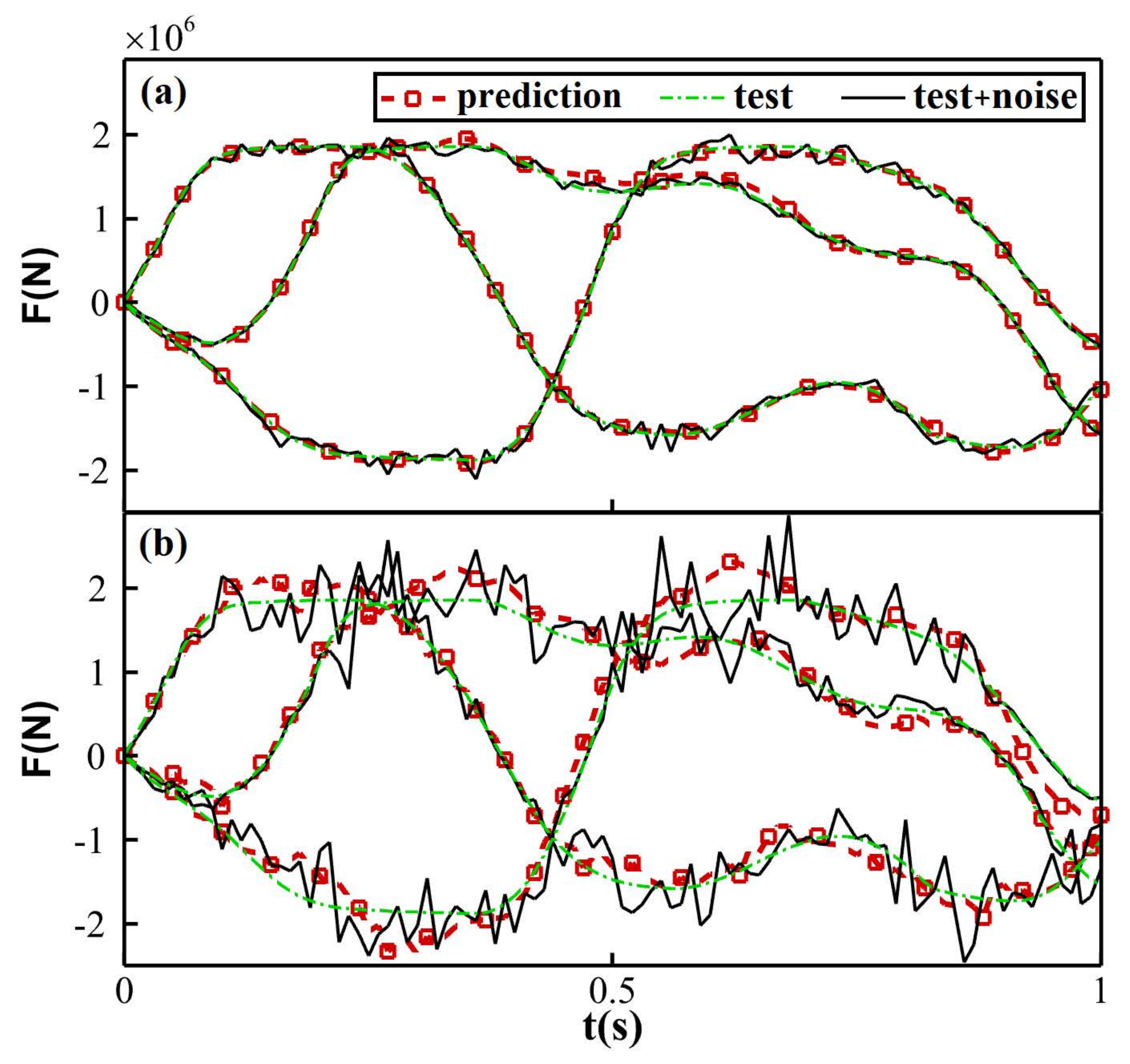}
    \caption{\textbf{Prediction of output data with noise.} (a) The prediction of output with 5\% noise. (b) The prediction of output with 20\% noise.}
    \label{fig13}
\end{figure}

The robustness of the neural network model could be critical if the training data is obtained from experimental data with ineliminable noise. To test the robustness of the DNO model, we add Gaussian white noise to the output data by
\begin{equation}\label{eq:noise}
  \widetilde{\sigma}=\sigma(1+\beta\xi),
\end{equation}
where $\xi$ is a Gaussian random variable with zero mean and unit variance, and $\beta\in[0.05,0.2]$ is an adjustable parameter setting the noise level. Subsequently, we train the DNO model based on the contaminated data. Fig.~\ref{fig13}(a) and Fig.~\ref{fig13}(b) present the results of $\beta=0.05$ and $\beta=0.2$, respectively, where we compare the predicted output, the test output, and the test output with noise. It is observed from Fig.~\ref{fig13} that the prediction results are smoother and more similar to the original noise-free test data compared with the test data with noise, which indicates that the trained DNO model is not overfitted.

Based on the testing of DNO trained on data with noise, the average predicted accuracy is $\chi=96.25\%$ for $\beta=0.05$, $\chi=91.70\%$ for $\beta=0.1$, $\chi=89.97\%$ for $\beta=0.15$, and $\chi=86.48\%$ for $\beta=0.2$. It means that when facing contaminated data, the DNO model can still provide good prediction accuracy better than $90\%$ as long as the noise level $\beta\le0.15$.


\section{Conclusion}\label{sec4}
We have developed a DNO model to act as a surrogate of physical models of an IPC beam to provide a fast and accurate prediction of transient mechanical responses of the IPC beam subject to dynamic loading.
The challenge of learning the time-dependent relationship between sequences can be solved by training a DNO model. Unlike FEA-based methods that require duplicated calculation when facing different dynamic loadings, the trained DNO model can process arbitrary inputs of dynamic loading and make predictions of the mechanical response of the nonlinear beam system instantly.

Since all the discretized elements within the IPC beam obey the same physical laws, this allows the DNO model to directly train one element on top of another. In the case of the same training data, by introducing an incremental learning algorithm to the DNO training, the prediction accuracy can be improved and the training process can be significantly accelerated. When the DNO model is trained using the incremental learning algorithm, the order of training also has a certain impact on the performance of the model. We noticed that adjacent elements have better similarity, and if training is performed sequentially according to the similarity of elements, better training results can be obtained.

We have compared the wall time for training a DNO model and training a traditional DeepONet model and found that the DNO model implemented in JAX supported by sequence-to-sequence training and incremental learning can be trained two orders of magnitude faster than the original DeepONet model implemented in TensorFlow.
The generality and extensibility of the DNO model have been tested. Given the DNO model is trained based on the Gaussian random process inputs, three different types of inputs, i.e., similar random Gaussian process input with longer time, sinusoidal, and piecewise linear sequences, are utilized as test cases to study the prediction accuracy of the DNO model. Results suggested that the DNO model still has good prediction accuracy for different types of inputs. Moreover, the DNO model also has good extensibility. Though the DNO model was trained on one second of data, it provided good prediction results when the test cases are with an extended time of 2 seconds dynamic strain loadings. In general, after offline training, the DNO model can act as a surrogate of physics-based FEA to predict the transient mechanical response instantly in terms of reaction force and stress distribution of the IPCs to various strain loads. Such fast and accurate prediction of the mechanical properties of IPCs could significantly accelerate the IPC structural design and related composite designs for desired mechanical properties.

The DNO model also shows good robustness. We have added Gaussian white noise at different noise level into the original output data to form new training data sets. Obviously, as the noise level increases, the prediction accuracy gradually decreases. In the case of $\beta=0.2$, the average prediction accuracy of the DNO model can still reach $\chi=86.48\%$, indicating great robustness. However, with a further increase in noise level, the structure of the data will also be polluted. To seek more reliable prediction results for noise data, other recent methods such as Bayesian neural network models~\cite{2021-Yang-BPINN,2022-Meng-leaning}, would be interesting research directions to explore in the future.

\section*{Acknowledgments}
This research is primarily supported as part of the AIM for Composites, an Energy Frontier Research Center funded by the U.S. Department of Energy (DOE), Office of Science, Basic Energy Sciences (BES), under Award \#DE-SC0023389 and by the National Science Foundation (Grants OAC-2103967 and CDS\&E-2204011).
We also acknowledge research support from Clemson University with generous allotment of computation time on the Palmetto cluster.

\bibliography{References}

\begin{thebibliography}{10}

\bibitem{2015-Murr-Examples}
Lawrence~E. Murr.
\newblock {\em Examples of Natural Composites and Composite Structures}.
\newblock Springer International Publishing, Cham, 2015.

\bibitem{2022-Tamura-Origin}
Yuto Tamura, Marie Tani, and Rei Kurita.
\newblock Origin of nonlinear force distributions in a composite system.
\newblock {\em Scientific Reports}, 12(1):632, 2022.

\bibitem{2014-Zheng-Ultralight}
Xiaoyu Zheng, Howon Lee, Todd~H. Weisgraber, Maxim Shusteff, Joshua DeOtte,
  Eric~B. Duoss, Joshua~D. Kuntz, Monika~M. Biener, Qi~Ge, Julie~A. Jackson,
  Sergei~O. Kucheyev, Nicholas~X. Fang, and Christopher~M. Spadaccini.
\newblock Ultralight, ultrastiff mechanical metamaterials.
\newblock {\em Science}, 344(6190):1373--1377, 2014.

\bibitem{2015-Clausen-Topology}
Anders Clausen, Fengwen Wang, Jakob~S. Jensen, Ole Sigmund, and Jennifer~A.
  Lewis.
\newblock Topology optimized architectures with programmable {Poisson's} ratio
  over large deformations.
\newblock {\em Advanced Materials}, 27(37):5523--5527, 2015.

\bibitem{2009-Aoyanagi-Stress}
Yuko Aoyanagi and Ko~Okumura.
\newblock Stress and displacement around a crack in layered network systems
  mimicking nacre.
\newblock {\em Phys. Rev. E}, 79:066108, 2009.

\bibitem{2020-Solala-OnThe}
Iina Solala, Maria~C. Iglesias, and Maria~S. Peresin.
\newblock On the potential of lignin-containing cellulose nanofibrils {LCNFs}:
  a review on properties and applications.
\newblock {\em Cellulose}, 27(4):1853--1877, 2020.

\bibitem{2021-Ghazlan-Inspiration}
Abdallah Ghazlan, Tuan Ngo, Ping Tan, Yi~Min Xie, Phuong Tran, and Matthew
  Donough.
\newblock Inspiration from nature's body armours – a review of biological and
  bioinspired composites.
\newblock {\em Composites Part B: Engineering}, 205:108513, 2021.

\bibitem{2022-Tiwary-AReview}
Akash Tiwary, Raman Kumar, and Jasgurpreet~Singh Chohan.
\newblock A review on characteristics of composite and advanced materials used
  for aerospace applications.
\newblock {\em Materials Today: Proceedings}, 51(1):865--870, 2022.

\bibitem{2022-Ma-Strength}
Guowei Ma, Wenwei Yang, and Li~Wang.
\newblock Strength-constrained simultaneous optimization of topology and fiber
  orientation of fiber-reinforced composite structures for additive
  manufacturing.
\newblock {\em Advances in Structural Engineering}, 25(7):1636--1651, 2022.

\bibitem{2018-Brenken-Fused}
Bastian Brenken, Eduardo Barocio, Anthony Favaloro, Vlastimil Kunc, and
  R.~Byron Pipes.
\newblock Fused filament fabrication of fiber-reinforced polymers: A review.
\newblock {\em Additive Manufacturing}, 21:1--16, 2018.

\bibitem{2018-Moustafa-Mesostructure}
Abdel~R. Moustafa, Ralph~B. Dinwiddie, Alexander~E. Pawlowski, Derek~A.
  Splitter, Amit Shyam, and Zachary~C. Cordero.
\newblock Mesostructure and porosity effects on the thermal conductivity of
  additively manufactured interpenetrating phase composites.
\newblock {\em Additive Manufacturing}, 22:223--229, 2018.

\bibitem{2016-Thompson-design}
Mary~Kathryn Thompson, Giovanni Moroni, Tom Vaneker, Georges Fadel, R.~Ian
  Campbell, Ian Gibson, Alain Bernard, Joachim Schulz, Patricia Graf, Bhrigu
  Ahuja, and Filomeno Martina.
\newblock Design for additive manufacturing: Trends, opportunities,
  considerations, and constraints.
\newblock {\em CIRP Annals}, 65(2):737--760, 2016.

\bibitem{1996-W-microstr}
W.~Liu and U.~Köster.
\newblock Microstructures and properties of interpenetrating alumina/aluminium
  composites made by reaction of sio2 glass preforms with molten aluminium.
\newblock {\em Materials Science and Engineering: A}, 210(1):1--7, 1996.

\bibitem{2003-San-alumina}
C.~San~Marchi, M.~Kouzeli, R.~Rao, J.~A. Lewis, and D.~C. Dunand.
\newblock Alumina–aluminum interpenetrating-phase composites with
  three-dimensional periodic architecture.
\newblock {\em Scripta Materialia}, 49(9):861--866, 2003.

\bibitem{2019-Goh-Recent}
Guo~Dong Goh, Yee~Ling Yap, Shweta Agarwala, and Wai~Yee Yeong.
\newblock Recent progress in additive manufacturing of fiber reinforced polymer
  composite.
\newblock {\em Advanced Materials Technologies}, 4(1):1800271, 2019.

\bibitem{2000-LDW-mechanical1}
L.D. Wegner and L.J. Gibson.
\newblock The mechanical behaviour of interpenetrating phase composites – i:
  modelling.
\newblock {\em International Journal of Mechanical Sciences}, 42(5):925--942,
  2000.

\bibitem{2019-Bonatti-mechanical}
Colin Bonatti and Dirk Mohr.
\newblock Mechanical performance of additively-manufactured anisotropic and
  isotropic smooth shell-lattice materials: Simulations \& experiments.
\newblock {\em J. Mech. Phys. Solids}, 122:1--26, 2019.

\bibitem{2017-Palaganas-3D}
N.~B. Palaganas, J.~D. Mangadlao, A.~C.~C. de~Leon, J.~O. Palaganas, K.~D.
  Pangilinan, Y.~J. Lee, and R.~C. Advincula.
\newblock 3d printing of photocurable cellulose nanocrystal composite for
  fabrication of complex architectures via stereolithography.
\newblock {\em ACS Appl Mater Interfaces}, 9(39):34314--34324, 2017.

\bibitem{2017-Wang-3D}
Xin Wang, Man Jiang, Zuowan Zhou, Jihua Gou, and David Hui.
\newblock 3d printing of polymer matrix composites: A review and prospective.
\newblock {\em Composites Part B: Engineering}, 110:442--458, 2017.

\bibitem{2021-Zhang-mechanical}
Yunfei Zhang, Meng-Ting Hsieh, and Lorenzo Valdevit.
\newblock Mechanical performance of 3d printed interpenetrating phase
  composites with spinodal topologies.
\newblock {\em Composite Structures}, 263, 2021.

\bibitem{1995-Helge-strength}
Helge Prielipp, Mathias Knechtel, Nils Claussen, S.K. Streiffer, H.~Müllejans,
  M.~Rühle, and Jürgen Rödel.
\newblock Strength and fracture toughness of aluminum/alumina composites with
  interpenetrating networks.
\newblock {\em Materials Science and Engineering: A}, 197(1):19--30, 1995.

\bibitem{2008-Poniznik-effective}
Z.~Poniznik, V.~Salit, M.~Basista, and D.~Gross.
\newblock Effective elastic properties of interpenetrating phase composites.
\newblock {\em Computational Materials Science}, 44(2):813--820, 2008.

\bibitem{2014-Cheng-modeling}
Feifei Cheng, Sun-Myung Kim, J.~N. Reddy, and Rashid~K. Abu Al-Rub.
\newblock Modeling of elastoplastic behavior of stainless-steel/bronze
  interpenetrating phase composites with damage evolution.
\newblock {\em International Journal of Plasticity}, 61:94--111, 2014.

\bibitem{2017-Al-mechanical}
Oraib Al-Ketan, Mhd Adel~Assad, and Rashid~K. Abu Al-Rub.
\newblock Mechanical properties of periodic interpenetrating phase composites
  with novel architected microstructures.
\newblock {\em Composite Structures}, 176:9--19, 2017.

\bibitem{2009-Binner-processing}
Jon Binner, Hong Chang, and Rebecca Higginson.
\newblock Processing of ceramic-metal interpenetrating composites.
\newblock {\em Journal of the European Ceramic Society}, 29(5):837--842, 2009.

\bibitem{2015-Liu-cyclic}
Shaobo Liu, Aiqun Li, Siyuan He, and Peng Xuan.
\newblock Cyclic compression behavior and energy dissipation of aluminum
  foam–polyurethane interpenetrating phase composites.
\newblock {\em Composites Part A: Applied Science and Manufacturing},
  78:35--41, 2015.

\bibitem{2017-Okulov-dealloying}
I.~V. Okulov, J.~Weissmuller, and J.~Markmann.
\newblock Dealloying-based interpenetrating-phase nanocomposites matching the
  elastic behavior of human bone.
\newblock {\em Sci Rep}, 7(1):20, 2017.

\bibitem{2019-Liu-mechanical}
Shaobo Liu, Aiqun Li, and Peng Xuan.
\newblock Mechanical behavior of aluminum foam/polyurethane interpenetrating
  phase composites under monotonic and cyclic compression.
\newblock {\em Composites Part A: Applied Science and Manufacturing},
  116:87--97, 2019.

\bibitem{2009-Jhaver-processing}
Rahul Jhaver and Hareesh Tippur.
\newblock Processing, compression response and finite element modeling of
  syntactic foam based interpenetrating phase composite (ipc).
\newblock {\em Materials Science and Engineering: A}, 499(1-2):507--517, 2009.

\bibitem{2014-Li-simulation}
Guoju Li, Xu~Zhang, Qunbo Fan, Linlin Wang, Hongmei Zhang, Fuchi Wang, and
  Yangwei Wang.
\newblock Simulation of damage and failure processes of interpenetrating sic/al
  composites subjected to dynamic compressive loading.
\newblock {\em Acta Materialia}, 78:190--202, 2014.

\bibitem{2014-Wang-damage}
Fu-chi Wang, Xu~Zhang, Yang-wei wang, Qun-bo Fan, and Guo-ju Li.
\newblock Damage evolution and distribution of interpenetrating phase
  composites under dynamic loading.
\newblock {\em Ceramics International}, 40(8):13241--13248, 2014.

\bibitem{2000-LDW-mechanical2}
L.D. Wegner and L.J. Gibson.
\newblock The mechanical behaviour of interpenetrating phase composites – ii:
  a case study of a three-dimensionally printed material.
\newblock {\em International Journal of Mechanical Sciences}, 42(5):943--964,
  2000.

\bibitem{2022-Li-ceramic}
Xinwei Li, Minseo Kim, and Wei Zhai.
\newblock Ceramic microlattice and epoxy interpenetrating phase composites with
  simultaneous high specific strength and specific energy absorption.
\newblock {\em Materials \& Design}, 223, 2022.

\bibitem{2022-Tong-experiment}
Lewei Tong, Luhua Chen, Xiaoqing Wang, Jia Zhu, Xiaodong Shao, and Zheng Zhao.
\newblock Experiment and finite element analysis of bending behavior of high
  strength steel-uhpc composite beams.
\newblock {\em Engineering Structures}, 266, 2022.

\bibitem{2021-Guo-constitutive}
Qian Guo, Wenjin Yao, Wenbin Li, and Nikhil Gupta.
\newblock Constitutive models for the structural analysis of composite
  materials for the finite element analysis: A review of recent practices.
\newblock {\em Composite Structures}, 260, 2021.

\bibitem{2021-Metin-in-plane}
Fatih Metin and Ahmet Avcı.
\newblock In-plane quasi-static and out-of-plane dynamic behavior of nanofiber
  interleaved glass/epoxy composites and finite element simulation.
\newblock {\em Composite Structures}, 270, 2021.

\bibitem{2021-Lin-Operator}
Chensen Lin, Zhen Li, Lu~Lu, Shengze Cai, Martin Maxey, and George~Em
  Karniadakis.
\newblock Operator learning for predicting multiscale bubble growth dynamics.
\newblock {\em The Journal of Chemical Physics}, 154(10):104118, 2021.

\bibitem{2022-Malidarre-invest}
Roya~Boodaghi Malidarre, Iskender Akkurt, Parisa~Boodaghi Malidarreh, and Seher
  Arslankaya.
\newblock Investigation and ann-based prediction of the radiation shielding,
  structural and mechanical properties of the hydroxyapatite (hap)
  bio-composite as artificial bone.
\newblock {\em Radiation Physics and Chemistry}, 197, 2022.

\bibitem{2022-Cai-applic}
Ruijun Cai, Kui Wang, Wei Wen, Yong Peng, Majid Baniassadi, and Said Ahzi.
\newblock Application of machine learning methods on dynamic strength analysis
  for additive manufactured polypropylene-based composites.
\newblock {\em Polymer Testing}, 110, 2022.

\bibitem{2022-Yin-interf}
Minglang Yin, Enrui Zhang, Yue Yu, and George~Em Karniadakis.
\newblock Interfacing finite elements with deep neural operators for fast
  multiscale modeling of mechanics problems.
\newblock {\em Computer Methods in Applied Mechanics and Engineering}, 2022.

\bibitem{2022-Tao-finite}
Fei Tao, Xin Liu, Haodong Du, and Wenbin Yu.
\newblock Finite element coupled positive definite deep neural networks
  mechanics system for constitutive modeling of composites.
\newblock {\em Computer Methods in Applied Mechanics and Engineering}, 391,
  2022.

\bibitem{2021-kovachki-neural}
Nikola Kovachki, Zongyi Li, Burigede Liu, Kamyar Azizzadenesheli, Kaushik
  Bhattacharya, Andrew Stuart, and Anima Anandkumar.
\newblock Neural operator: Learning maps between function spaces.
\newblock {\em arXiv preprint arXiv:2108.08481}, 2021.

\bibitem{2020-Li-Fourier}
Zongyi Li, Nikola Kovachki, Kamyar Azizzadenesheli, Burigede Liu, Kaushik
  Bhattacharya, Andrew Stuart, and Anima Anandkumar.
\newblock Fourier neural operator for parametric partial differential
  equations, 2020.

\bibitem{2021-Lu-learning}
Lu~Lu, Pengzhan Jin, Guofei Pang, Zhongqiang Zhang, and George~Em Karniadakis.
\newblock Learning nonlinear operators via deeponet based on the universal
  approximation theorem of operators.
\newblock {\em Nature Machine Intelligence}, 3(3):218--229, 2021.

\bibitem{1995-Asko-design}
Asko Talja and Pekka Salmi.
\newblock {\em Design of stainless steel RHS beams, columns and beam-columns}.
\newblock Number 1619 in VTT Tiedotteita - Meddelanden - Research Notes. VTT
  Technical Research Centre of Finland, 1995.

\bibitem{2003-Kim-fullrange}
Kim~J.R. Rasmussen.
\newblock Full-range stress–strain curves for stainless steel alloys.
\newblock {\em Journal of Constructional Steel Research}, 59(1):47--61, 2003.

\bibitem{2003-Yu-experi}
Yufeng Zha and Torgeir Moan.
\newblock Experimental and numerical prediction of collapse of flatbar
  stiffeners in aluminum panels.
\newblock {\em Journal of Structural Engineering}, 129(2):160--168, 2003.

\bibitem{2021-Yun-fullrange}
Xiang Yun, Zhongxing Wang, and Leroy Gardner.
\newblock Full-range stress-strain curves for aluminum alloys.
\newblock {\em Journal of Structural Engineering}, 147(6), 2021.

\bibitem{2016-Loyala-smart}
R.~Dg Loyola, M.~Pedergnana, and S.~Gimeno~Garcia.
\newblock Smart sampling and incremental function learning for very large high
  dimensional data.
\newblock {\em Neural Netw}, 78:75--87, 2016.

\bibitem{2021-Lu-DeepXDE}
Lu~Lu, Xuhui Meng, Zhiping Mao, and George~Em Karniadakis.
\newblock Deepxde: A deep learning library for solving differential equations.
\newblock {\em SIAM Review}, 63(1):208--228, 2021.

\bibitem{2018-James-JAX}
James Bradbury, Roy Frostig, Peter Hawkins, Matthew~James Johnson, Chris Leary,
  Dougal Maclaurin, George Necula, Adam Paszke, Jake Vander{P}las, Skye
  Wanderman-{M}ilne, and Qiao Zhang.
\newblock {JAX}: composable transformations of {P}ython+{N}um{P}y programs,
  2018.

\bibitem{2021-Yang-BPINN}
Liu Yang, Xuhui Meng, and George~Em Karniadakis.
\newblock {B-PINNs: Bayesian} physics-informed neural networks for forward and
  inverse {PDE} problems with noisy data.
\newblock {\em Journal of Computational Physics}, 425:109913, 2021.

\bibitem{2022-Meng-leaning}
Xuhui Meng, Liu Yang, Zhiping Mao, José del~Águila Ferrandis, and George~Em
  Karniadakis.
\newblock Learning functional priors and posteriors from data and physics.
\newblock {\em J. Comput. Phys.}, 457, 2022.

\end{thebibliography}

\end{document}